\documentclass[journal]{IEEEtran}
\usepackage{cite}
\usepackage{multirow}
\usepackage[table,xcdraw]{xcolor}
\usepackage{amsmath}        
\usepackage{amssymb}        
\usepackage{algorithm}      
\usepackage{algpseudocode}
\usepackage{array}
\usepackage[caption=false,font=normalsize,labelfont=sf,textfont=sf]{subfig}
\usepackage{textcomp}
\usepackage{stfloats}
\usepackage{url}
\usepackage{verbatim}
\usepackage{graphicx}
\usepackage{color}
\begin{document}

\title{ZISVFM: Zero-Shot Object Instance Segmentation in Indoor Robotic Environments with Vision Foundation Models}

\author{\IEEEauthorblockN{Ying~Zhang,~\IEEEmembership{Senior Member,~IEEE,} Maoliang~Yin, Wenfu~Bi, Haibao~Yan, \\ Shaohan Bian, Cui-Hua Zhang,~\IEEEmembership{Member,~IEEE}, and Changchun Hua,~\IEEEmembership{Fellow,~IEEE}
                          }
\thanks{This work was supported in part by the National Natural Science Foundation of China under Grant 62203378, 62203377, 62073279,
in part by the Hebei Natural Science Foundation under Grant No. F2024203036, F2024203115, F2024203051, in part by the Science Research Project of Hebei Education Departmen under Grant BJK2024195, and in part by the S\&T Program of Hebei under Grant 236Z2002G, 236Z1603G. \emph{(Corresponding author: Cui-Hua Zhang.)}}

\thanks{The authors are with the School of Electrical Engineering, Yanshan University, Qinhuangdao, 066004, China. (e-mail: yzhang@ysu.edu.cn; mlyin@stumail.ysu.edu.cn; wenfubi@gmail.com; hby97@stumail.ysu.edu.cn; bsh@stumail.ysu.edu.cn; cuihuazhang@ysu.edu.cn; cch@ysu.edu.cn).

{\color{blue}This article has been accepted for publication in IEEE Transactions on Robotics. This is the author's version which has not been fully edited and content may change prior to final publication.

DOI: 10.1109/TRO.2025.3539198}

1941-0468 © 2024 IEEE. Personal use is permitted, but republication/redistribution requires IEEE permission.}
}
\markboth{This article has been accepted for publication in a future issue of \textbf{\emph{IEEE Transactions on Robotics}}.}%
{Shell \MakeLowercase{\textit{et al.}}: IEEE Transactions on Robotics}

\maketitle

\begin{abstract}
Service robots operating in unstructured environments must effectively recognize and segment unknown objects to enhance their functionality. Traditional supervised learning-based segmentation techniques require extensive annotated datasets, which are impractical for the diversity of objects encountered in real-world scenarios. Unseen Object Instance Segmentation (UOIS) methods aim to address this by training models on synthetic data to generalize to novel objects, but they often suffer from the simulation-to-reality gap. This paper proposes a novel approach (ZISVFM) for solving UOIS by leveraging the powerful zero-shot capability of the segment anything model (SAM) and explicit visual representations from a self-supervised vision transformer (ViT). 
The proposed framework operates in three stages: (1) generating object-agnostic mask proposals from colorized depth images using SAM, (2) refining these proposals using attention-based features from the self-supervised ViT to filter non-object masks, and (3) applying K-Medoids clustering to generate point prompts that guide SAM towards precise object segmentation. Experimental validation on two benchmark datasets and a self-collected dataset demonstrates the superior performance of ZISVFM in complex environments, including hierarchical settings such as cabinets, drawers, and handheld objects. Our source code is available at https://github.com/Yinmlmaoliang/zisvfm.
\end{abstract}


\section{Introduction}
For service robots operating in unstructured environments like homes, hospitals, or supermarkets, the capability to recognize and segment novel objects is paramount \cite{xie2021unseen, fu2024taylor, richtsfeld2012segmentation}. Given the vast number of objects a robot might encounter in its working environment, it is impractical to model each one. Instead, the ability to segment unseen objects allows robots to handle, grasp, and manipulate unfamiliar objects and tools, significantly enhancing their functionality across diverse real-world settings \cite{ceola2022learn, xiang2021learning, li2023stow}. This critical capability can lead to more efficient and adaptable robotic applications in both domestic and industrial environments. Therefore, in this work, our aim is to enable the robot in a zero-shot manner to segment arbitrary objects of interest in various environmental settings, such as tabletop, handheld, etc.

Traditional supervised learning-based segmentation techniques \cite{he2017mask,minaee2021image} depend significantly on extensive human-generated annotations found in datasets like ImageNet \cite{russakovsky2015imagenet} and MS COCO \cite{lin2014microsoft}. However, there is a notable absence of large-scale, realistic datasets with a sufficient variety of objects for robot manipulation scenarios. Furthermore, creating such datasets that reflect the diversity of objects in human environments is impractical.
To tackle this issue, recent research efforts \cite{back2022unseen,xiang2021learning,9636281} have focused on Unseen Object Instance Segmentation (UOIS), where the methods use synthetic data for category-agnostic training aimed at enabling the model to develop a concept of ``\emph{objectness}" and generalize it to novel objects.
Some of these methods \cite{xie2020best, danielczuk2019segmenting} combine the effective generalization capabilities of depth images with the ability of RGB images to generate clear masks, achieving good segmentation performance in the real world.
Nevertheless, for robotic perception, synthetic data often fail to accurately represent various real-world aspects, such as object texture, lighting conditions, and depth noise. This simulation-to-reality (sim2real) gap significantly hampers the model's performance on realistic data, particularly in novel environments. Further advancements involve the application of domain adaptation techniques \cite{zhang2023unseen}, which enhance the model's efficacy by bridging the gap between synthetic training environments and real-world conditions. Additionally, interaction perception based UOIS approaches \cite{lu2023self, yu2022self, eitel2019self} implement strategies that actively acquire labels for unknown objects in real application environments through robot actions, such as grasping and pushing. These methods leverage the newly acquired real-world data to fine-tune a pre-trained model, enhancing segmentation performance in authentic robotic settings. However, these methods either rely on synthetic data for training, which often suffers from sim2real gap, or require some post-processing, which can be inflexible during robot operation.

Recent advancements in foundational models for image segmentation, such as the segment anything model (SAM), have shown significant progress across various applications \cite{kirillov2023segment,wang2023seggpt,chen2023semantic,zou2023segment}. Developed using the SA-1B dataset, which includes 11 million images and over 1 billion masks, SAM exhibits exceptional zero-shot generalization abilities. It efficiently generates precise object masks from various prompts, including bounding boxes and specific points. However, SAM's main limitation lies in its inability to consistently capture complete instances, often leading to over-segmentation.

In this work, we propose a novel approach named ZISVFM, which exploits the zero-shot generalization and feature representation capabilities of vision foundation models to address the UOIS challenge in robot perception without additional training. The methodology involves three main stages. First, we generate object-agnostic mask proposals by converting depth images to the RGB color space using the viridis color map \cite{hunter2007matplotlib}. These colorized depth images are then input into SAM with a low non-maximum suppression (NMS) threshold, generating a diverse set of object-agnostic mask proposals that include a mixture of foreground objects, background regions, and noise masks. In the second stage, we focus on the removal of non-object masks using a strategy that leverages explicit visual representations from a vision transformer (ViT) trained with DINOv2 \cite{oquab2023dinov2}. The explicit visual representations are observable, semantically meaningful patterns that arise in the ViT's attention maps as a result of self-supervised learning. We extract the attention maps from ViT with the feature representation, where the attention maps highlight the objects of interest regions, and weight the features according to the information entropy of the attention maps to prioritise the more attentive head features. Then, we use the image patch that receives the least attention as the background patch, and calculate a similarity matrix between the image patch and the background patch using the weighted features to identify and remove non-object masks. Finally, in the third stage, the refined set of object masks serves as proposal areas for object locations. Within each area, positive points are sampled using K-Medoids clustering and are used as prompts for SAM to perform the final object segmentation, enhancing the accuracy of object masks, particularly in edge regions. The proposed methodology effectively leverages the strengths of SAM and explicit visual representations from a self-supervised ViT to accurately segment objects in images. The proposed ZISVFM outperforms baseline methods on two benchmark datasets and achieves comparable performance to some state-of-the-art (SOTA) methods that require training. In addition, we utilize a fetch robot to collect a dataset from complex, multi-level environments including cabinets, drawers, desktops, and handheld objects, extending beyond the single-plane limitation. On this challenging dataset, ZISVFM significantly outperforms the benchmark approaches. Further grasping experiments on unknown objects in real-world settings validate the practicality of the proposed method.

The main contributions of this paper are as follows:
\begin{enumerate} 
  \item A novel framework, ZISVFM, is proposed, which leverages the exceptional zero-shot segmentation ability of SAM and the explicit visual representations from a self-supervised ViT to accurately segment unknown objects in real-world robotic environments across diverse scenarios, including hierarchical settings.
  \item We show that the more attentively focused features of ViT heads can be effectively emphasized by integrating a weighting strategy based on the information entropy of attention maps, enabling robust non-object mask removal through background similarity analysis.
  \item The effectiveness and robustness of ZISVFM are demonstrated through extensive experiments on both benchmark datasets and a self-collected dataset featuring complex, multi-level environments. The results show ZISVFM exhibits superior performance compared to baseline methods and achieves comparable performance to SOTA methods.
  \item The robotic application of ZISVFM is demonstrated, highlighting its capability to facilitate the grasping of unknown objects in cluttered environments\footnotemark[1]\footnotetext[1]{https://sites.google.com/view/zisvfmzero}.
\end{enumerate}

The rest of this paper is structured as follows: Section \ref{section2} introduces a brief description of the related works. Section \ref{section3} presents the ZISVFM methodology in detail. Section \ref{section4} discusses the experimental setup, datasets, and evaluation metrics employed to assess the performance of ZISVFM, along with the experimental results. Finally, Section \ref{section5} concludes this paper.

\section{Related Work} \label{section2}
In this section, an overview of methods for UOIS is first outlined, which mainly focuses on application in the robotics. Then, a brief overview of the vision foundation models is presented. 
\subsection{Unseen Objects Instance Segmentation}
Recent advances in UOIS have significantly enhanced robotic capabilities in unstructured environments. The primary challenge in this field is the scarcity of large-scale, real-world datasets typically required for robotic applications. To overcome this, recent approaches utilize synthetic data to train UOIS models. Xie \emph{et al}. \cite{xie2020best, xie2021unseen} advocate for the use of synthetic RGB and depth data in a two-stage process: initially generating coarse initial masks from depth data alone, which are subsequently refined with RGB data. Despite the synthetic nature of RGB-D data, this approach has proven effective in producing precise and detailed masks. On the other hand, Xiang \emph{et al.} \cite{xiang2021learning} introduce a technique for learning RGB-D feature embeddings from synthetic data, enabling the application of a mean shift clustering algorithm to segment previously unseen objects. For the Unseen Object Amodal Instance Segmentation (UOAIS) task, Back \emph{et al}. \cite{back2022unseen} design UOAIS-Net to highlight the importance of amodal perception for robotic manipulation in cluttered environments. In order to address the sim-to-real gap, Zhang \emph{et al}. \cite{zhang2023unseen} propose the Fully Test-time RGB-D Embeddings Adaptation (FTEA) framework, which employs test-time domain adaptation to enhance real-world segmentation performance. Instead of depth input, the Instance Stereo Transformer (INSTR) \cite{9636281} uses stereo image pairs and a transformer-based architecture to segment objects directly. Recent strategies aim to enhance segmentation under challenging conditions such as densely packed or overlapping objects by further refining the results of the UOIS model. For example, RICE \cite{xie2022rice} addresses occlusions in cluttered scenes by employing a graph-based representation to refine the results of the UOIS model. While models trained on synthetic data benefit from domain transfer techniques \cite{zhang2023unseen}, the sim-to-real gap still introduces errors in real-world applications. Recently, Lu \emph{et al}. \cite{lu2023self} and Yu \emph{et al}. \cite{yu2022self} introduce a novel robotic system for improving the UOIS performance in the real world by leveraging long-term robot interaction with objects, utilizing the robot's maneuvers (e.g., pushing, grasping) to generate labeled datasets. Subsequently, fine-tuning the segmentation network with this real-world data markedly boosts accuracy both within and across different domains.

However, the aforementioned methods either rely on synthetic training data, which often results in a significant sim2real gap that hampers real-world performance, or necessitate extensive post-processing steps, which can be cumbersome and inflexible during dynamic robot operation. Advancements in computer vision have facilitated the development of methods \cite{chen20233d, nguyen2023cnos} that utilize computer-aided design (CAD) models for segmenting objects, leveraging the zero-shot capabilities of vision foundation models like SAM \cite{kirillov2023segment}. These methods use SAM to create a series of mask proposals, matched against CAD model references to segment objects at the instance level without prior training. Nevertheless, acquiring a CAD model for each object is nearly impossible in practice.

\subsection{Vision Foundation Model}
Recent advancements have driven the development of foundation models, which are trained on extensive, diverse datasets and can be fine-tuned for various related downstream tasks. In this section, we provide a brief overview of the vision foundation models relevant to our work, mainly oriented towards segmentation and self-supervised vision models.
\subsubsection{Segment Anything Model (SAM)}
There are several types of image segmentation model, including panoptic \cite{kirillov2019panoptic}, instance \cite{bolya2019yolact}, and semantic segmentation \cite{long2015fully}. Current segmentation models are specialized according to the segmentation type or dataset used. Foundational segmentation models strive to develop universally applicable solutions for various segmentation tasks \cite{awais2023foundational}. An exemplar of this approach is SAM \cite{kirillov2023segment}, a zero-shot segmentation model trained with 1.1 billion masks and only 11 million images, demonstrating remarkable segmentation capabilities. SAM not only segments images autonomously but also utilizes input prompts such as foreground/background points or bounding boxes to enhance segmentation performance. Thanks to its robust capabilities, SAM has been applied to a range of computer vision tasks, including medical image segmentation \cite{he2023accuracy}, object tracking \cite{cheng2023segment}, and image style transfer \cite{yu2023inpaint}. However, recent studies \cite{cui2023all,zhang2023uvosam} suggest that SAM may underperform in domain-specific tasks, especially when provided with limited input prompts, occasionally failing to segment specific object instances. In this work, we address this limitation by supplying SAM with precise object point prompts to accurately segment the target object in the image.
\subsubsection{Self-supervised Vision Transformers}
The successful application of transformer architectures \cite{vaswani2017attention} to vision-related tasks, as exemplified by the ViT \cite{dosovitskiy2020image}, has underscored the potential of these models as potent image encoders for supervised tasks \cite{zhang2023cross}. This capability extends into the self-supervised domain, as demonstrated by MoCo-v3\cite{chen2021mocov3}, which employs ViT for self-supervised representation learning through contrastive learning techniques, yielding impressive results.
Additionally, DINO \cite{caron2021emerging} leverages knowledge distillation alongside a momentum encoder and multi-crop training to explore the local-to-global correspondence within the vision transformer. This reveals ViT's inherent capacity to handle complex tasks such as semantic segmentation.
To further expand the utility of ViT in self-supervised learning, MST \cite{li2021mst} integrates concepts from the BERT architecture \cite{devlin2018bert} by dynamically masking and then predicting the masked tokens in an image, using a global decoder. Similarly, BEIT \cite{bao2021beit}  proposes a masked image modeling task to recover the original visual tokens based on the corrupted image patches. Another approach, masked autoencoders (MAE) \cite{he2022masked}, involves masking substantial portions of the input image and learning to reconstruct the omitted pixels, enhancing the model's predictive and generative capabilities.
Recent studies \cite{amir2021deep,caron2021emerging,oquab2023dinov2} have demonstrated that self-attention maps in Vision Transformers encode localization information critical for detecting and segmenting salient objects in images. Our research leverages these self-supervised features of ViT to effectively differentiate foreground from background in scene images, illustrating their practical utility in complex vision tasks.
\begin{figure*}
	\centering
	\includegraphics[width=6in]{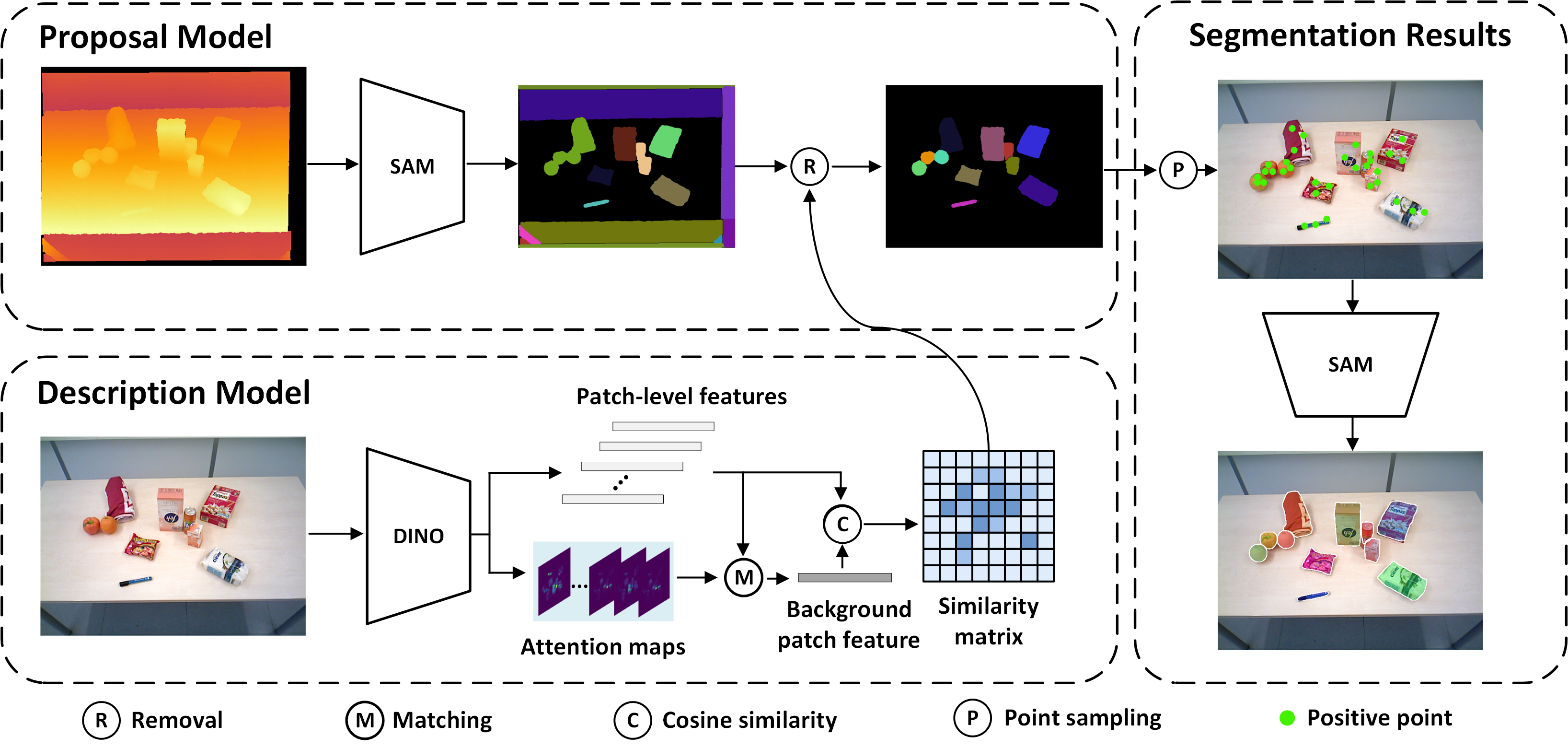}
	\centering
	\caption{Overview of the proposed ZISVFM methodology. This approach employs two vision foundation models: SAM \cite{kirillov2023segment} for segmentation and ViT trained with DINOv2 \cite{oquab2023dinov2} for feature description in a scene. The process consists of three main stages: 1) Generating object-agnostic mask proposals using SAM on colorized depth images; 2) Refinement of object masks by removing non-object masks based on  explicit visual representations from a self-supervised ViT; 3) Point prompts derived from clustering centres within each object's proposal further optimise object segmentation performance.}
	\label{fig1}
\end{figure*}

\section{Method} \label{section3}
\subsection{Preliminary} \label{section31}
\subsubsection{Revisit of SAM}
SAM \cite{kirillov2023segment} is a zero-shot segmentation model capable of identifying and segmenting any object within an image without specific training on those objects. This capability stems from its training on the expansive SA-1B dataset, which encompasses over one billion segmentation masks from a wide variety of images. SAM operates in two primary modes: the Generator and the Predictor. The Generator mode segments all objects within an image but does not support targeted segmentation. In contrast, the Predictor mode offers enhanced precision by allowing the segmentation of specific objects based on prompts, such as bounding boxes or points of interest.

The architecture of SAM includes three principal components: the image encoder, the prompt encoder, and the mask decoder. The image encoder processes images of 1024 $\times$ 1024 resolution to produce a compact 64 $\times$ 64 spatial size image embedding. The prompt encoder processes various prompt types, from sparse prompts like points, boxes, and text, to dense prompts such as masks, converting them into c-dimensional tokens. The mask decoder then merges the image and prompt embeddings to effectively predict segmentation masks.
\subsubsection{Revisit of DINO}
DINO \cite{caron2021emerging} is a self-supervised learning framework designed for ViT to enhance feature extraction capabilities in visual models without requiring labeled data. DINO employs knowledge distillation within a teacher-student framework, where the student model learns by emulating the output of a more advanced teacher model. This framework utilizes two networks, identical in architecture but distinct in weight configurations. The student is trained using backpropagation, while the teacher updates its weights through an exponential moving average of both its own and the student's weights. During training, the networks see different and randomly transformed input parts. The student network predicts the mean-centered output of the teacher, fostering an understanding of local image patches within the global context of the teacher model. Notably, the self-supervised training yields ViT features that distinctly capture the scene layout, including object boundaries.

Building on the original framework, DINOv2 \cite{oquab2023dinov2} delivers SOTA transfer performance through increased model size, dataset scale, and various training enhancements. For the purposes of this paper, the advancements and emergent features of DINOv2 are central to extracting and manipulating visual components, such as background removal.

\subsection{The overall of proposed ZISFVM} \label{section32}
This paper presents a novel approach for solving UOIS by leveraging the powerful zero-shot capability of SAM and explicit visual representations from a self-supervised ViT trained with DINOv2. The overall framework of the proposed methodology is illustrated in Fig. \ref{fig1}. The methodology consists of three main stages:

\begin{itemize}
    \item \textbf{Generation of object-agnostic mask proposals:} Depth images are first converted to the RGB color space using the viridis color map \cite{hunter2007matplotlib} to emphasize geometric information. These colorized depth images are then input into SAM with a low NMS threshold to generate a set of object agnostic mask proposals. The proposals include a mixture of foreground objects, background regions, and noise masks.
    \item \textbf{Removal of non-object masks:} Removal of non-object masks in our approach leverages a strategy based on explicit visual representations to select salient objects amidst a blend of foreground and background masks. Specifically, this process employs explicit visual representations through a ViT trained with DINOv2 to extract attention maps and feature representations. The attention maps highlight object regions and are weighted based on information entropy to emphasize heads with more focused attention. A background similarity matrix is computed using the weighted features to identify and remove non-object masks.
   \item \textbf{Segmentation with point prompts:} The refined set of object masks serve as proposal areas for object locations. Within each area, positive points are sampled using K-Medoids clustering and used as prompts for SAM to perform final object segmentation. This improves the accuracy of object masks, especially in edge regions.
\end{itemize}

\subsection{Generation of object-agnostic mask proposals} \label{section33}
\subsubsection{Pre-processing depth images} Contrasting with RGB images, a notable characteristic of depth images is the pronounced emphasis on geometric information, accentuating the contours of objects over aspects like texture and color \cite{zhang2023cross}. As shown in Fig. \ref{fig1}, the input image of the model intuitively demonstrates this phenomenon. To exploit the inherent properties of depth maps, our approach converts these maps to the RGB color space. In this work, we exclusively use viridis color maps. After conversion, these depth maps are used as inputs to SAM. This process helps to extract complete object mask proposals independent of color and texture.
\subsubsection{Segmentation with SAM} Given a colorized depth image $I_d \in \mathbb{R} ^{H \times W}$, we utilize the SAM with default settings, except for setting a low NMS threshold of 0.5, to generate a set of $N_m$ object-agnostic mask proposals, $M=\left\{m_{1}, m_{2}, ..., m_{N_m}\right\}$, where $m_{i} \in \left\{0, 1 \right\}^{H \times W}$ for $i = 1, 2, ..., N_m$. It is worthy to note that $N_m$ is not fixed and varies depending on the content of each $I_d$. As illustrated in Fig. \ref{fig1}, the segmentation results $M$ includes the mixture of foreground objects and background masks, as well as noise masks originating from the depth image. Within the object masks, due to the lack of texture and color information in the depth images, cases of object occlusion or stacking result in combinations of object masks in the segmentation results. However, it is noteworthy that, despite such special cases, the segmentation result $M$ still contains independent masks for each object. To ensure that each segmentation mask $m_i$ represents an independent region without overlap, we present a post-processing algorithm to the masks generated by the SAM automatic segmentation mode. The pseudocode for this algorithm is detailed in Algorithm \ref{alg:make_masks_independent}. The algorithm operates in two main stages. First, it identifies and removes ``union masks'', which are masks that can be approximated by the union of other smaller masks. For each mask, we consider all combinations of up to \( k_{\text{max}} \) other masks and compute the Intersection over Union (IoU) between the mask in question and the union of these combinations. If the IoU exceeds a threshold \( \theta \), the mask is deemed redundant and is removed from further consideration.
In the second stage, we sort the remaining masks by their area in ascending order to prioritize processing of smaller masks first. We then iterate over each mask and remove any regions that overlap with previously processed masks. This procedure ensures that the final set of masks $M'=\{m'_1, m'_2, \dots, m'_{N_{m'}}\}$ is independent, with each mask uniquely representing a distinct region in the image without overlap.
\begin{algorithm}
    \caption{Make Masks Independent}
    \label{alg:make_masks_independent}
    \begin{algorithmic}[1]
    \Require
        Set of binary masks $M=\{m_1, m_2, \dots, m_{N_m}\}$ of size $H \times W$, intersection over union (IoU) threshold $\theta$ (default: 0.8), maximum combination size $k_{\text{max}}$ (default: 3).
    \Ensure
        Set of independent masks $M'=\{m'_1, m'_2, \dots, m'_{N_{m'}}\}$
    \Statex
    \Comment{\textbf{Part 1: Remove Union Masks}}
    \State Ensure all masks are in binary format
    \State Compute area $A_i$ of each mask $m_i$ by summing pixels
    \State Initialize empty set $U$ to store indices of union masks
    \For{$i = 1$ to $N_m$}
        \If{$i \in U$}
            \State \textbf{continue}
        \EndIf
        \For{$k = 2$ to $k_{\text{max}}$}
            \State $S \gets \{ j \mid j \neq i,\, j \notin U \}$
            \If{$|S| < k$}
                \State \textbf{break}
            \EndIf
            \For{each combination $C \subset S$ where $|C| = k$}
                \State Compute union mask $U_C = \bigcup_{j \in C} m_j$
                \State Compute IoU:
                \[
                    \text{IoU} = \frac{|m_i \cap U_C|}{|m_i \cup U_C|}
                \]
                \If{$\text{IoU} \geq \theta$}
                    \State Add $i$ to $U$
                    \State \textbf{break}
                \EndIf
            \EndFor
            \If{$i \in U$}
                \State \textbf{break}
            \EndIf
        \EndFor
    \EndFor
    \State Remove masks $m_i$ where $i \in U$
    \Statex
    \Comment{\textbf{Part 2: Process Overlapping Masks}}
    \State Sort remaining masks $\{m_i\}$ in ascending order of area $A_i$
    \State Initialize combined mask $C = \mathbf{0}_{H \times W}$
    \State Initialize empty list \textit{FinalMasks}
    \For{each mask $m_i$ in sorted order}
        \State Compute independent mask $m'_i = m_i \setminus C$
        \If{$m'_i$ contains any pixels}
            \State Add $m'_i$ to \textit{FinalMasks}
            \State Update combined mask $C = C \cup m'_i$
        \EndIf
    \EndFor
    \State \Return \textit{FinalMasks}
    \end{algorithmic}
\end{algorithm}
\subsection{Removal of non-object masks} \label{section34}
To select salient objects from the blend of foreground and background masks $M'$ generated by SAM, we propose a strategy based on explicit visual representations derived from DINOv2. This approach employs a ViT trained with DINOv2, extracting latent variables from its final layer. The explicit visual representations of objects contained in these variables are utilized to effectively distinguish and retain object masks while removing non-object masks in $M'$. This method leverages the rich semantic understanding capabilities of DINOv2, enabling robust object selection regardless of the object's size or position in the image.

Given an RGB image $I_r \in \mathbb{R} ^{H \times W \times 3}$, ViT \cite{dosovitskiy2020image} trained with DINOv2 \cite{oquab2023dinov2} takes non-overlapping 2D image patches of resolutions $K\times K$ as inputs, with the number of patches $N_p=HW/K^2$. Each patch is represented as a token, described by a vector of numerical features. An extra learnable token, denoted as a class token \texttt{CLS}, is used to represent the aggregated information of the entire set of patches. To remove background masks in $M'$, the latent features of the image are first extracted from the final layer of the ViT. 
In the multi-head self-attention mechanism of the final layer, there are $N_h$ different heads, each capturing distinct visual information. For each head $i$, we extract the attention map $a_i \in \mathbb{R}^{N_p \times N_p}$ representing the attention weights between \texttt{CLS} token and all image patch tokens. The key feature representations from the final layer are denoted as $F\in \mathbb{R} ^{N_h \times N_p \times d}$, where $d$ represents the feature dimension per head.

\begin{figure}
	\includegraphics[width=2.5in]{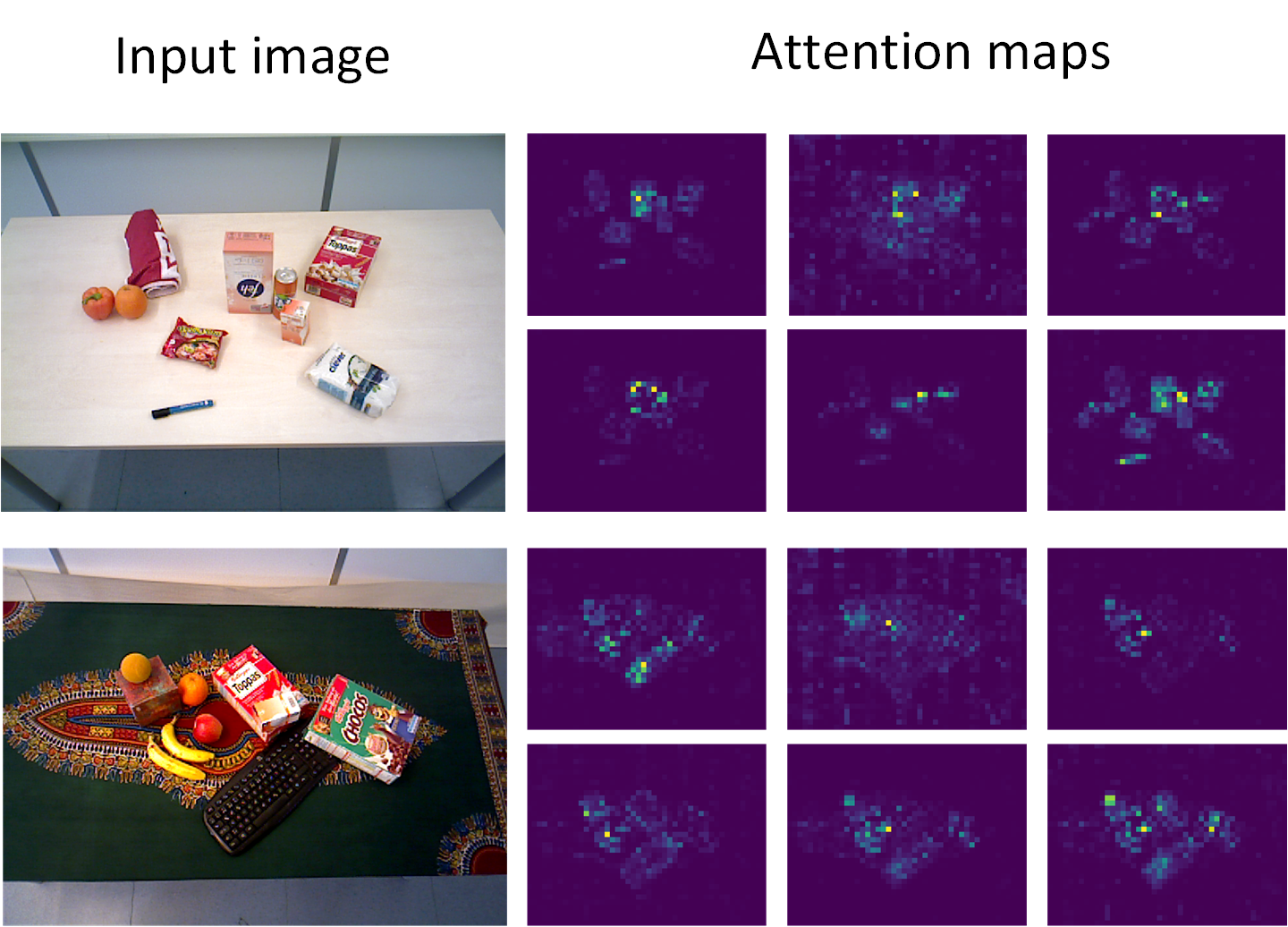}
	\centering
	\caption{Visualization of self-attention maps obtained with the six different heads in the last attention layer from ViT.}
	\label{fig3}
\end{figure}

ViT employs an attention mechanism, where the attention maps are a visual representation of these attention values. 
Fig. \ref{fig3} visualizes the attention maps $a_i$ (for $i = 1, 2, ... N_p$) generated by the ViT-S/14 model (a smaller variant with a 14$\times$14 patch size) trained on the DINOv2 framework \cite{oquab2023dinov2} and applied to images from the OCID dataset \cite{suchi2019easylabel}.  
As shown in Fig. \ref{fig3}, these attention patterns tend to concentrate on semantically meaningful regions, while showing relatively lower activation in background areas. This characteristic can be easily exploited to identify which masks in $M$ belong to the background. However, it is noteworthy that the background appears more or less clearly in different heads, with attention in these heads appearing more dispersed. Therefore, to accurately determine which masks in $M'$ belong to the background, we first perform weighting on the attention map $a_i$ and features $F$ of different heads based on the information entropy of the attention map. By assigning greater weight to the heads with focused attention, the positional information of objects is emphasized. Finally, the background patches are identified by leveraging the explicit visual representations contained in the weighted attention maps, which pertain to the images. The weighted features then contribute to the computation of a background similarity matrix, essential for eliminating background masks from $M'$.

Formally, we commence by calculating the weights for each attention map $a_i$, based on the concept of information entropy. Information entropy, a statistical measure of uncertainty, quantifies the informational content in attention maps. For each attention map $a_i$, its information entropy $E\left ( a_{i}  \right )$ is computed, and it is subsequently utilized to determine the weights. The entropy is calculated using the formula:
\begin{equation}
	E\left ( a_{i}   \right )  = - { \sum^{N_p}_{p=1}P(a_{pi})\log_{2}{P(a_{pi})}}
\end{equation}
where $P(\cdot)$ represents normalization operation for each attention map $a_i$, $a_{pi}$ is the attention value for patch $p$ from head $i$. Following this, the weighting formula is as follows:
\begin{equation}
	\omega _i = -\log_{}{\left ( \frac{E (a_{i} )}{\textstyle \sum_{j=1}^{N_h}E (a_{j} ) } \right ) }
\end{equation}
This ensures that heads with lower entropy (more focused attention) receive higher weights $w_i$.
As a result of the above calculations, the final vector of weights $W=[\omega_{1}, \omega_{2}, ..., \omega_{N_h}]$ is obtained.

Upon obtaining the vector of weights, a similarity matrix between a background patch and all other image patches is calculated to determine whether a patch belongs to the background. To compute this similarity matrix, a background patch is first defined as the patch with the minimum sum of activation values in the attention map $a_i$. Subsequently, the similarity matrix is determined by using the cosine similarity between patch-level features. Specifically, the index of the background patch is obtained by the following formula:
\begin{equation}
	l =\underset{p\in\left \{ 1, ..., N_p \right \}  }{argmin}\sum_{i}^{N_h} \omega _i a_{pi}
\end{equation}
Then, the weighted features $F_\omega$ is given by $F_\omega=W\odot F$, where ``$\odot$'' represents element-wise multiplication. Subsequently, $F_\omega$ is reshaped from $\mathbb{R} ^{N_h\times N_p \times d}$ into $\mathbb{R} ^{N_p\times D}$, where $D=N_h \times d$ represents the embedding dimension of each image patch. The cosine similarity matrix $\mathcal{S}_{cos} \in \mathbb{R} ^{h \times w}$ is calculated as follows:

\begin{equation}
	F_{norm }^{'} = \frac{ F^{'} }{\sqrt{ {\textstyle \sum_{j=1}^{D}F_{pj}^{2} } } } , \; p=1,2,...,N_p
\end{equation}
\begin{equation}
	\mathcal{S}_{cos}=reshape\left ( F_{norm}^{'} \cdot F_{norm}^{'}\left [ l, : \right ], h, w\right )
\end{equation}
Here, $h=H/K$ and $w=W/K$ represent the height and width of the feature map, respectively. The values in the background similarity matrix describe the score for an image patch to be considered as a background patch. Consequently, we average the scores corresponding to the image patches of masks in $M'$, assigning a score to each mask. When this score exceeds a predetermined threshold $\tau $ , the mask is considered as background and is thus removed.

\subsection{Segmentation with point prompts} \label{section35}
After removing non-object masks from $M'$, the collection of object masks are obtained, denoted as $M_o$. Due to the quality limitations of depth images, the accuracy of object masks in edge regions is relatively low, which may affect the effectiveness of subsequent tasks performed by robots. Therefore, the object masks obtained from depth images are used as proposal areas for object locations. Within these proposal areas, we sample a set of positive points as prompts and employ SAM for segmentation. Specifically, given an RGB image $I_r \in \mathbb{R} ^{H \times W \times 3}$, the proposal areas $M_o$ for object locations are obtaind from its corresponding depth image. In each area, the cluster centers from the K-Medoids clustering algorithm \cite{park2009simple} are used as positive points. This method ensures good coverage of different parts of the object and robustness to noise and outliers, as demonstrated by the ablation experiments in the experimental section. Finally, these point prompts are adopted to guide SAM in performing segmentation, achieving improved segmentation results, as shown in the Fig. \ref{fig1}.
\section{Experiments} \label{section4}
In this section, we elaborate the experimental details of the proposed ZISVFM and perform performance evaluation on 
three real-world datasets (OCID, OSD, and  self-collected HIOD datasets) to validate the effectiveness. Also, the performance comparison of our model with baseline and SOTA methods is investigated. Besides, a series of ablation studies are conducted to demonstrate the contribution of each component of our approach. Then, examples of common failures of the proposed method and the reasons are discussed. Finally, the effectiveness and usefulness of our proposed method are evaluated in a real-world setting using a Fetch mobile robot. The supplementary video is available at https://sites.google.com/view/zisvfmzero.

\subsection{Implementation Details} \label{section41}
In the proposed approach, we employ two primary vision foundation models: SAM \cite{kirillov2023segment}, as a segmentation model, and ViT, trained with DINOv2 \cite{oquab2023dinov2}, as a feature description model for scene representation. Throughout the experiments, the SAM model incorporates a ViT-H/16 backbone \cite{dosovitskiy2020image}, whereas the description model utilizes a ViT-B/14 \cite{dosovitskiy2020image} architecture pre-trained with \cite{darcet2023vitneedreg}. The SAM model's parameters are carefully configured to ensure optimal segmentation performance. The box non-maximum suppression threshold ($\texttt{box\_nms\_thresh} = 0.5$) controls the overlap tolerance between predicted bounding boxes, where higher values allow more overlapping predictions to be retained. The crop overlap ratio ($\texttt{crop\_overlap\_ratio} = 0$) determines the extent of overlap between adjacent image crops during processing, with zero indicating no overlap between successive crops. The minimum mask region area ($\texttt{min\_mask\_region\_area} = 0$) specifies the smallest acceptable size for a segmented region, where a value of zero allows the detection of segments of any size. The points per batch parameter ($\texttt{points\_per\_batch} = 64$) defines the number of prompt points processed simultaneously. The stability score threshold ($\texttt{stability\_score\_thresh} = 0.95$) sets a high confidence requirement for segment prediction, ensuring that only highly reliable segmentations are retained in the final output, where values closer to 1 indicate more stringent filtering of predictions.

The mask filtering phase incorporates three key parameters: the IoU threshold ($\theta = 0.8$), a maximum combination size ($3$), and a background similarity threshold ($\tau = 0.47$). In the final stage, we apply the K-Medoids clustering algorithm to each object proposal to generate three point prompts. All experiments are conducted within the PyTorch \cite{paszke2019pytorch} framework using an NVIDIA RTX A6000 graphics card.

\subsection{Datasets and Evaluation Metrics} \label{section42}
The model's performance is evaluated on three distinct datasets: the Object Clutter Indoor Dataset (OCID) \cite{suchi2019easylabel}, the Object Segmentation Database (OSD) \cite{richtsfeld2012segmentation}, and our newly introduced Hierarchical Indoor Object Dataset (HIOD). The OCID \cite{suchi2019easylabel} comprises 2,390 RGB-D images, each featuring up to 20 distinct objects, with an average of 3.3 objects per image. It is obtained through a semi-automatic annotation process, where objects are incrementally added to scenes and labels are created by evaluating depth differences. Such an approach, while facilitating the annotation of a large number of images, introduces a degree of noise due to the variability inherent in depth measurements. The OSD dataset consists of 111 RGB-D images, each containing up to 15 objects and averaging 3.3 objects per image. It is distinguished by its manual annotation process, which ensures high precision and reliability in the segmentation masks.

To address the limitations of existing benchmarks that primarily focus on single-plane scenarios, we introduce HIOD, a dataset specifically designed to evaluate object detection in complex, hierarchical indoor environments. HIOD encompasses challenging multi-layered configurations commonly encountered in real-world robotics applications, including objects placed in cabinets, drawers, and across different planar layers, as well as handheld objects. The dataset is structured into four principal scene categories: cabinet arrangements, drawer configurations, sofa-desktop combinations, and handheld object scenarios. Data collection is conducted using a Fetch robot, with viewpoint selection carefully determined based on the robot's degrees of freedom within each scene type. For instance, drawer scenes are captured from five strategic viewpoints: frontal, bilateral, and two oblique angles. The HIOD dataset comprises 74 sets of depth images with corresponding RGB images, each annotated with manually created segmentation masks to ensure precise boundary delineation. This meticulous annotation approach guarantees high-quality ground truth data for robust model evaluation.

Following\cite{Dave_2019_ICCV}, we evaluate object segmentation performance using precision, recall, and F-measure. Specifically, the F-measure is computed between all pairs of predicted and ground truth masks, which are matched via the Hungarian method. Consequently, precision, recall, and F-measure can be represented as:
\begin{equation*}
P=\frac{\sum_{i}|p_{i}\cap g(p_{i})|}{\sum_{i}|p_{i}|},R=\frac{\sum_{i}|p_{i}\cap g(p_{i})|}{\sum_{j}|g_{j}|},F=\frac{2P R}{P+R}
\end{equation*}
where $p_i$ denotes the segmentation of predicted object $i$, $g(p_i)$ corresponds to the segmentation of the ground truth object matched with $p_i$, and $g_j$ represents the segmentation of ground truth object $j$. The Overlap $P/R/F$ indicates the aforementioned metrics. Additionally, Boundary $P/R/F$ metrics are used to evaluate the accuracy of predicted boundaries against ground truth boundaries, with true positives determined by the pixel overlap of these boundaries. Furthermore, $F@.75$ represents the percentage of segmented objects with an Overlap F-measure greater than 75$\%$.

\subsection{Quantitative Results} \label{section43}

\begin{table*}[]
    \centering
    \caption{The UOIS performances of the proposed method on OCID\cite{suchi2019easylabel} and OSD\cite{richtsfeld2012segmentation} datasets compared to the baseline method. ``-'' denotes the segmentation results are not refined using point prompts.}
    \begin{tabular}{l|l|ccccccc|ccccccc}
        \hline
                                 &                         & \multicolumn{7}{c|}{OCID\cite{suchi2019easylabel}}                                                                                                                                                                                                                & \multicolumn{7}{c}{OSD\cite{richtsfeld2012segmentation}}                                                                                                                                                                                                                  \\ \cline{3-16}
                                 &                         & \multicolumn{3}{c|}{Overlap}                                                                        & \multicolumn{3}{c|}{Boundary}                                                                       & {\color[HTML]{000000} }      & \multicolumn{3}{c|}{Overlap}                                                                        & \multicolumn{3}{c|}{Boundary}                                                                       & {\color[HTML]{000000} }      \\
        \multirow{-3}{*}{Method} & \multirow{-3}{*}{Input} & {\color[HTML]{CE6301} P} & {\color[HTML]{329A9D} R} & \multicolumn{1}{c|}{{\color[HTML]{9A0000} F}} & {\color[HTML]{CE6301} P} & {\color[HTML]{329A9D} R} & \multicolumn{1}{c|}{{\color[HTML]{9A0000} F}} & {\color[HTML]{CD9934} F@.75} & {\color[HTML]{CE6301} P} & {\color[HTML]{329A9D} R} & \multicolumn{1}{c|}{{\color[HTML]{9A0000} F}} & {\color[HTML]{CE6301} P} & {\color[HTML]{329A9D} R} & \multicolumn{1}{c|}{{\color[HTML]{9A0000} F}} & {\color[HTML]{CD9934} F@.75} \\ \hline
        SAM\cite{kirillov2023segment}                      & RGB                     & 27.9                     & 70.8                     & \multicolumn{1}{c|}{40.0}                     & 24.0                     & 78.1                     & \multicolumn{1}{c|}{36.7}                     & 63.7                         & 30.2                     & 76.4                     & \multicolumn{1}{c|}{43.3}                     & 29.4                     & \textbf{78.8}            & \multicolumn{1}{c|}{42.7}                     & \textbf{70.6}                \\
        SAM\cite{kirillov2023segment}                      & Depth                   & 82.7                     & \textbf{86.4}            & \multicolumn{1}{c|}{84.4}                     & 76.7                     & 81.0                     & \multicolumn{1}{c|}{78.5}                     & 83.6                         & 51.3                     & \textbf{76.9}            & \multicolumn{1}{c|}{60.0}                     & 36.3                     & 53.1                     & \multicolumn{1}{c|}{41.3}                     & 63.6                         \\
        \textbf{ZISVFM}-         & RGBD                    & 90.8           & 85.7                     & \multicolumn{1}{c|}{88.2}            & 82.4            & 81.5            & \multicolumn{1}{c|}{82.0}            & 86.9                & 79.9           & 73.3                     & \multicolumn{1}{c|}{76.5}            & 52.5            & 53.8                     & \multicolumn{1}{c|}{53.1}            & 66.0                    \\
        \textbf{ZISVFM}         & RGBD                    & \textbf{92.5}           & 86.1                    & \multicolumn{1}{c|}{\textbf{89.2}}            & \textbf{84.5}            & \textbf{82.4}            & \multicolumn{1}{c|}{\textbf{83.5}}            & \textbf{87.0}               & \textbf{86.4}            & 73.1                    & \multicolumn{1}{c|}{\textbf{78.1}}            & \textbf{61.7}           & 58.5                     & \multicolumn{1}{c|}{\textbf{60.0}}            & 66.0    \\ \hline
        \end{tabular} \label{table1}
\end{table*}

\begin{table}[]
    \centering
    \caption{Compared the UOIS performance of the proposed method with the baseline method and other SOTA methods on our HIOD dataset. ``+'' denotes the zoom-in cluster refinement operation \cite{xiang2021learning} to refine segmentation results. ``-'' denotes the segmentation results are not refined using point prompts.}
    \resizebox{\linewidth}{!}{
        \begin{tabular}{l|l|ccccccc}
            \hline
            \multicolumn{1}{c|}{}                         & \multicolumn{1}{c|}{}                        & \multicolumn{7}{c}{HIOD}                                                                                                                                                                                                                                   \\ \cline{3-9}
            \multicolumn{1}{c|}{}                         & \multicolumn{1}{c|}{}                        & \multicolumn{3}{c|}{Overlap}                                                                        & \multicolumn{3}{c|}{Boundary}                                                                       & {\color{blue} }                        \\
            \multicolumn{1}{c|}{\multirow{-3}{*}{Method}} & \multicolumn{1}{c|}{\multirow{-3}{*}{Input}} & {P} & {R} & \multicolumn{1}{c|}{F} & {P} & {R} & \multicolumn{1}{c|}{F} & {F@.75} \\ \hline
            SAM\cite{kirillov2023segment}                  & RGB                                          & 21.5                     & 74.1                     & \multicolumn{1}{c|}{33.3}                     & 28.3                     & 78.4                     & \multicolumn{1}{c|}{41.5}                     & 52.6                                           \\
            SAM\cite{kirillov2023segment}                 & Depth                                        & 67.6                     & 81.1                     & \multicolumn{1}{c|}{73.7}                     & 52.7                     & 62.4                     & \multicolumn{1}{c|}{57.2}                     & 65.2                                           \\
            UOIS-Net-3D \cite{xie2021unseen}              & RGBD                                         & 72.8                     & 67.6                     & \multicolumn{1}{c|}{70.1}                     & 65.9                     & 64.1                     & \multicolumn{1}{c|}{65.0}                     & 71.5                                           \\
            MSMFormer+\cite{lu2022mean}                           & RGBD                                         & 78.1                     & 71.0                     & \multicolumn{1}{c|}{74.4}                     & 73.2                     & 65.7                     & \multicolumn{1}{c|}{69.2}                     & 78.7                                           \\
            ZISVFM-                                       & RGBD                                         & 89.4                     & 88.9                     & \multicolumn{1}{c|}{89.2}                     & 78.7                     & 76.8                     & \multicolumn{1}{c|}{77.8}                     & 87.0                                           \\
            ZISVFM                                        & RGBD                                         & \textbf{91.3}           & \textbf{89.7}           & \multicolumn{1}{c|}{\textbf{90.5}}           & \textbf{88.5}           & \textbf{85.9}           & \multicolumn{1}{c|}{\textbf{87.1}}           & \textbf{87.8}                                  \\ \hline
            \end{tabular}\label{table2}
    }
    \end{table}

\begin{table}[]
    \centering
    \caption{The UOIS performances of the proposed method and other SOTA methods on OCID\cite{suchi2019easylabel} dataset. ``+'' denotes the zoom-in cluster refinement operation \cite{xiang2021learning} to refine segmentation results. ``$*$'' indicates that the model with amodal perception}
    \resizebox{\linewidth}{!}{
        \begin{tabular}{l|l|llllllc}
            \hline
                                     & {\color[HTML]{000000} }                        & \multicolumn{7}{c}{OCID \cite{suchi2019easylabel}}                                                                                                                                                                                                                                                                                                                   \\ \cline{3-9}
                                     & {\color[HTML]{000000} }                        & \multicolumn{3}{c|}{Overlap}                                                                                                                & \multicolumn{3}{c|}{Boundary}                                                                                                               & {\color[HTML]{CD9934} }                        \\
            \multirow{-3}{*}{Method} & \multirow{-3}{*}{{\color[HTML]{000000} Input}} & \multicolumn{1}{c}{{\color[HTML]{CE6301} P}} & \multicolumn{1}{c}{{\color[HTML]{329A9D} R}} & \multicolumn{1}{c|}{{\color[HTML]{9A0000} F}} & \multicolumn{1}{c}{{\color[HTML]{CE6301} P}} & \multicolumn{1}{c}{{\color[HTML]{329A9D} R}} & \multicolumn{1}{c|}{{\color[HTML]{9A0000} F}} & \multirow{1}{*}{{\color[HTML]{CD9934} F@.75}} \\ \hline
            Mask R-CNN \cite{he2017mask}              & Depth                                          & \multicolumn{1}{c}{85.3}                     & \multicolumn{1}{c}{85.6}                     & \multicolumn{1}{c|}{84.7}                     & \multicolumn{1}{c}{83.2}                     & \multicolumn{1}{c}{76.6}                     & \multicolumn{1}{l|}{78.8}                     & 72.7                                           \\
            UOIS-Net-2D \cite{xie2020best}             & Depth                                          & \multicolumn{1}{c}{88.3}                     & 78.9                                         & \multicolumn{1}{l|}{81.7}                     & 82.0                                         & 65.9                                         & \multicolumn{1}{l|}{71.4}                     & 69.1                                           \\
            UOIS-Net-3D \cite{xie2021unseen}             & Depth                                          & 86.5                                         & 86.6                                         & \multicolumn{1}{l|}{86.4}                     & 80.0                                         & 73.4                                         & \multicolumn{1}{l|}{76.2}                     & 77.2                                           \\
            UOAIS-Net$*$ \cite{back2022unseen}               & Depth                                          & 89.9                                         & 90.9                                         & \multicolumn{1}{l|}{89.8}                     & 86.7                                         & 84.1                                         & \multicolumn{1}{l|}{84.7}                     & 87.1                                           \\
            UCN \cite{xiang2021learning}                     & RGBD                                           & 86.0                                         & 92.3                                         & \multicolumn{1}{l|}{88.5}                     & 80.4                                         & 78.3                                         & \multicolumn{1}{l|}{78.8}                     & 82.2                                           \\
            UCN+ \cite{xiang2021learning}                    & RGBD                                           & 91.6                                         & 92.5                                         & \multicolumn{1}{l|}{91.6}                     & 86.5                                         & 87.1                                         & \multicolumn{1}{l|}{86.1}                     & 89.3                                           \\
            FTEA \cite{zhang2023unseen}                    & RGBD                                           & 86.2                                         & 93.9                                         & \multicolumn{1}{l|}{89.5}                     & 79.5                                         & 79.5                                         & \multicolumn{1}{l|}{79.1}                     & 85.1                                           \\
            FTEA+ \cite{zhang2023unseen}                   & RGBD                                           & 92.0                                         & 93.3                                         & \multicolumn{1}{l|}{92.3}                     & 86.5                                         & 88.0                                         & \multicolumn{1}{l|}{86.7}                     & 91.1                                           \\
            MSMFormer \cite{lu2022mean}               & RGBD                                           & 88.4                                         & 90.2                                         & \multicolumn{1}{l|}{88.5}                     & 84.7                                         & 83.1                                         & \multicolumn{1}{l|}{83.0}                     & 80.3                                           \\
            MSMFormer+ \cite{lu2022mean}              & RGBD                                           & 92.5                                         & 91.0                                         & \multicolumn{1}{l|}{91.5}                     & 89.4                                         & 85.9                                         & \multicolumn{1}{l|}{87.3}                     & 86.0                                           \\
            \textbf{ZISVFM}            & RGBD                                           & 92.5                                         &  86.1                                        & \multicolumn{1}{l|}{89.2}                     & 84.5                                         & 82.4                                         & \multicolumn{1}{l|}{83.5}                     & 87.0                                           \\ \hline
            \end{tabular} \label{table3} }
\end{table}

\subsubsection{Comparison to Baselines}
In Table \ref{table1}, the performance of our method is compared against the baseline SAM \cite{kirillov2023segment} method on the OCID \cite{suchi2019easylabel} and OSD \cite{richtsfeld2012segmentation} datasets. SAM produces over-segmentation results in an automatic manner. To ensure a fairer comparison, both methods designated all predicted masks smaller than 500 pixels as background, and also set larger masks (those occupying more than 80\% of the entire image) as background. Since baseline methods aim to over-segment the scene, the precision is usually lower and the recall is higher. It is observed in Table \ref{table1} that when SAM takes depth images as input, it has the best performance for almost all metrics compared to taking RGB images as input, especially in terms of precision and F-measure, except for the boundary recall on the OSD dataset. This indicates SAM's sensitivity to color and texture variations, often resulting in incomplete object segmentation masks, whereas depth images are able to provide complete object masks. The boundary recall is very high on the OSD benchmark when SAM takes an RGB image as input, due to the fact that over-segmentation on the RGB image provides higher boundary quality compared to the depth image. The results on our HIOD dataset (Table \ref{table2}) further validate our method's effectiveness in handling complex hierarchical scenarios. Notably, our approach achieves significant improvements in precision (91.3\%) and F-measure (90.5\%) compared to the baseline SAM method, which achieves 67.6\% and 73.7\% respectively when using depth input.

Compared with the baseline SAM method, our method successfully combines the advantages of both RGB and depth modalities, resulting in more accurate and complete object segmentation across all datasets. This is particularly evident in the consistent improvement in F-measure and precision metrics, demonstrating robust performance in both standard benchmark scenarios and more challenging hierarchical environments.

\subsubsection{Comparison to SOTA}
A comprehensive comparison between our method and SOTA approaches is presented across both established benchmarks (Table \ref{table3}) and our challenging HIOD dataset (Table \ref{table2}). On the OCID dataset, our method achieves competitive results with 92.5\% precision, comparable to methods employing additional refinement operations. Notably, unlike other approaches that rely on large-scale synthetic training data such as the TOD dataset \cite{xie2021unseen}, our method operates without scene-specific training. Additionally, some methods employ zoom-in cluster refinement operation to further improve the segmentation results, which our method does not use. Except for the recall, our method demonstrates competitive precision and F-measure performance compared to methods that do not use refinement operations.

The results on our HIOD dataset (table \ref{table2}) demonstrate the superiority of the proposed approach in handling complex hierarchical scenarios. Our approach significantly outperforms both baseline and SOTA methods, achieving 91.3\% precision, 89.7\% recall, and 90.5\% F-measure. In contrast, established methods such as UOIS-Net-3D exhibit considerably lower performance (72.8\% precision, 67.6\% recall), while MSMFormer+ achieves intermediate results (78.1\% precision, 71.0\% recall).
A consistent pattern emerges across our experimental evaluation: our method exhibits superior precision while maintaining relatively lower recall compared to contemporary approaches. The robust object precision (91.3\% on HIOD, 92.5\% on OCID) indicates high reliability in object segment identification. This precision-recall trade-off can be attributed to three key factors:
\begin{itemize}
    \item \textbf{Depth-based Proposal Generation:} The initial object proposals are primarily derived from depth image information. In scenarios with minimal depth differentiation or objects in close proximity, the depth-based segmentation may fail to generate distinct proposals, resulting in reduced recall performance.
    \item \textbf{Visual Feature Refinement:} The incorporation of explicit visual representations from self-supervised ViT enables efficient background mask filtering during proposal generation, effectively minimizing false positives.
    \item \textbf{Attention Distribution Effect:} The self-supervised ViT's attention capacity becomes distributed across multiple regions as scene complexity increases. Consequently, objects with less salient features receive diminished attention weights, as evidenced by the performance metrics in Tables \ref{table2} and \ref{table3}, particularly in complex multi-object scenarios. 
\end{itemize}

\subsection{Qualitative Results}
\subsubsection{Comparison to Baselines}
We evaluate the ZISVFM method in comparison to baseline approaches (SAM-RGB and SAM-Depth) using both OCID and HIOD datasets (Fig. \ref{fig8}, first two columns). The baseline methods demonstrate distinct characteristics in their segmentation performance. Although SAM is initially developed for RGB images, our experimental results indicate that SAM-Depth generates more reliable object-centric mask proposals. This enhanced performance can be attributed to the fundamental properties of depth images relative to RGB inputs. In RGB-based segmentation, variations in texture, shadows, and illumination conditions frequently lead to oversegmentation, as SAM exhibits sensitivity to these appearance-based variations. This phenomenon is particularly evident in the OCID examples, where objects exhibiting complex textures or varying illumination conditions are frequently segmented into multiple distinct regions. Conversely, depth images provide more robust geometric cues that naturally correspond to object boundaries. The depth modality effectively mitigates the influence of surface appearance variations that often confound RGB-based segmentation, enabling SAM to focus predominantly on objects' geometric properties. These characteristics establish SAM-Depth mask proposals as a more reliable foundation for subsequent object-level segmentation refinement. However, it is important to note that this segmentation approach primarily relies on physical object boundaries rather than semantic category recognition. Consequently, while geometrically accurate, the segmentation remains semantically agnostic to the objects of interest within the scene, underscoring the necessity for incorporating semantic understanding in complex scene unknown object segmentation tasks.
\subsubsection{Comparison to SOTA}
\begin{figure*}
	\includegraphics[width=7in]{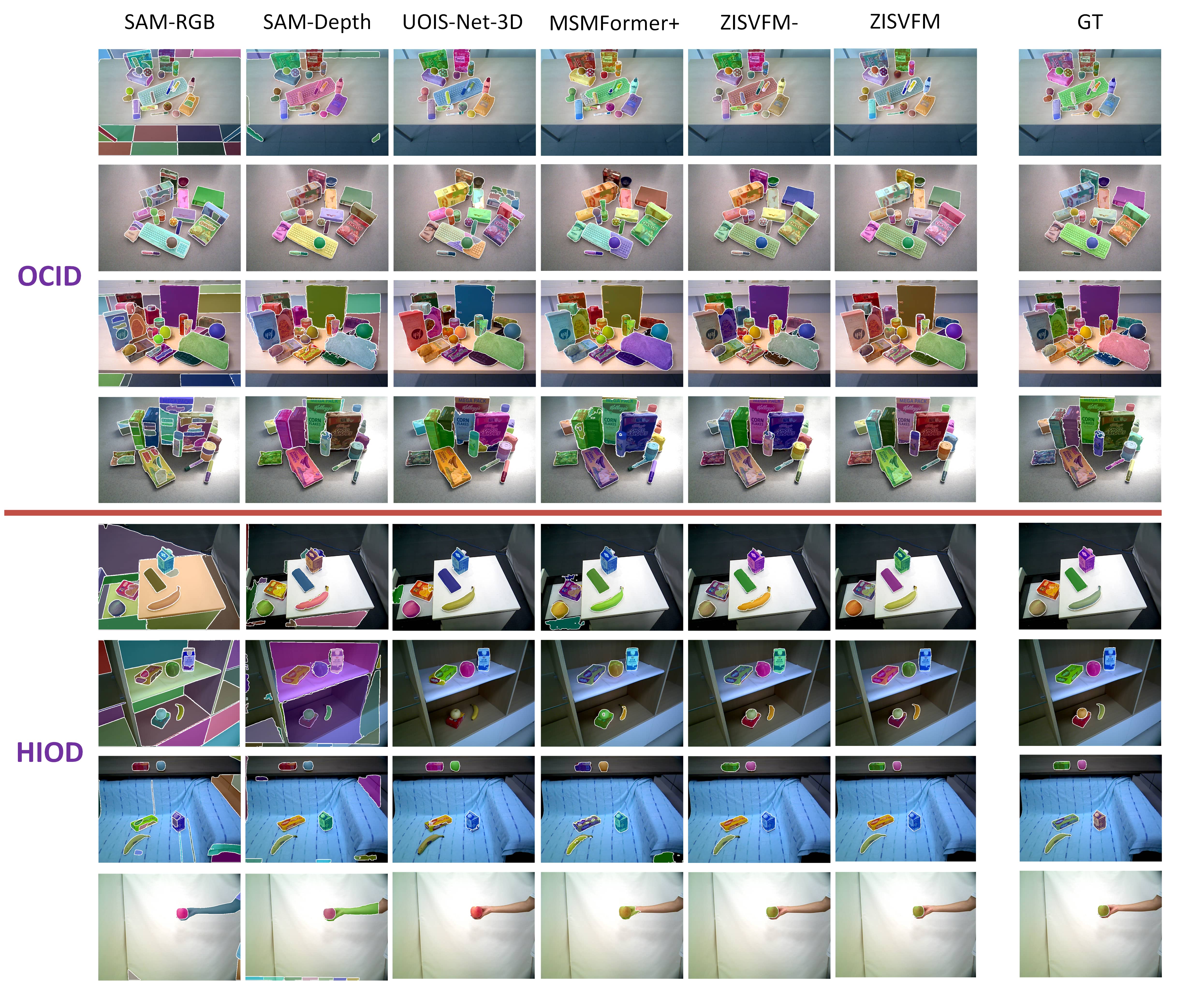}
	\centering
	\caption{Comparison of ZISVFM with baseline, SOTA methods on OCID and HIOD datasets. The baseline method, SAM, utilized RGB and depth images as inputs. The SOTA methods contain two representative UOIS methods, UOIS-Net-3D and MSMFormer, with the latter incorporating a zoom-in cluster refinement operation. In comparison to all baseline and advanced methods, our proposed ZISVFM demonstrated the capability to provide clear and precise masks in the hierarchical scenes of HIOD.}
	\label{fig8}
\end{figure*}

We further compare our method against SOTA approaches, including UOIS-Net-3D and MSMFormer. Both UOIS-Net-3D and MSMFormer are trained on the TOD dataset, which primarily comprises synthetic scenes with single-plane tabletop arrangements, and demonstrate varying performance across the two test datasets. On the OCID dataset, which closely mirrors the training domain's single-plane tabletop configurations (Fig. \ref{fig8}, top), both UOIS-Net-3D and MSMFormer demonstrate competent performance. This success can be attributed to the strong correlation between OCID's environmental setup and TOD's synthetic scenes, which feature ShapeNet objects arranged on convex tabletops. However, when evaluated on HIOD (Fig. \ref{fig8}, bottom), which introduces more complex, hierarchical scenarios, both methods exhibit significant performance degradation. HIOD's multi-layered configurations—including objects placed inside cabinets and drawers across different planar layers—present substantial challenges for maintaining consistent segmentation quality. As evidenced in the last row of Fig. \ref{fig8}, UOIS-Net-3D demonstrates particular limitations in handheld scene configurations. While MSMFormer achieves improved segmentation accuracy, it still exhibits confusion in object delineation, particularly in scenarios involving stacked objects within hierarchical scenes, as demonstrated in row 6.

In contrast, ZISVFM demonstrates robust performance across both datasets, with particularly notable advantages in HIOD's challenging scenarios. Our method successfully handles: (1) hierarchical spatial relationships, accurately segmenting objects across different levels such as cabinet shelves and drawer compartments, (2) complex geometric configurations that deviate significantly from simple tabletop arrangements, and (3) real-world robotic manipulation scenarios where objects exist in various spatial contexts. 
This superior generalization capability is evident in the HIOD examples, where ZISVFM maintains accurate segmentation across diverse scenarios including cabinet interiors, multi-level plane arrangements, and handheld object configurations, which pose significant challenges to existing SOTA methods due to their training data limitations. Moreover, Fig. \ref{fig8} also reveals that, leveraging SAM's robust segmentation capabilities, our method produces boundary delineation that surpasses ground truth accuracy when provided with object point prompts.

\subsection{Ablation Studies} \label{section44}
\subsubsection{Effective Weighting Strategy}
\begin{table}[]
    \centering
    \caption{Comparison of method overlap performance with and without weighting on OCID\cite{suchi2019easylabel} and OSD\cite{richtsfeld2012segmentation} datasets.}
    \begin{tabular}{l|ccc|ccc}
    \hline
                                & \multicolumn{3}{c|}{OCID \cite{suchi2019easylabel}}                                                      & \multicolumn{3}{c}{OSD \cite{richtsfeld2012segmentation}}                                                        \\
    \multirow{-2}{*}{Method} & {\color[HTML]{CE6301} P} & {\color[HTML]{329A9D} R} & {\color[HTML]{9A0000} F} & {\color[HTML]{CE6301} P} & {\color[HTML]{329A9D} R} & {\color[HTML]{9A0000} F} \\ \hline
    w/o                      & 88.9                     & 87.4                     & 87.9                    & 71.8                     & 72.1                    & 71.9                     \\
    w                        & 90.8                     & 85.7                     & 88.2                     & 79.9                     & 73.3                    & 76.5                     \\ \hline
    \end{tabular}
    \label{table4}
    \end{table}
Table \ref{table4} presents a comparative analysis of our proposed method's effectiveness with and without a weighting strategy on two datasets: OCID and OSD. The experimental results are not refined using point prompts due to the importance of the object position proposals generated from the second stage of the proposed method for accurately segmenting an instance. The incorporation of the weighting strategy results in enhanced precision for both datasets, which implies that it bolsters the method's capability to accurately identify instances while diminishing false positives. However, the recall slightly diminishes with the application of weighting. This may be due to the insufficient feature extraction capability of the ViT trained with DINOv2, which fails to capture the entirety of each object in the scene, thus causing the model to miss some real positive instances. The attention mechanism in ViT distributes attention weights across different regions of the image, and as the number of objects in the scene increases, the attention allocated to each object decreases, making their features less distinct. This phenomenon can be observed in the attention maps visualized from the images, where the attention is focused on a few prominent objects while other objects are neglected. Nevertheless, the F-measure for both datasets indicates an overall enhancement. Overall, the weighting strategy notably elevates model performance, particularly in terms of precision, suggesting it effectively optimizes model performance, maintaining high precision while reducing erroneous predictions.
\subsubsection{$\tau $ Ablation}
\begin{figure}
    \includegraphics[width=3.4in]{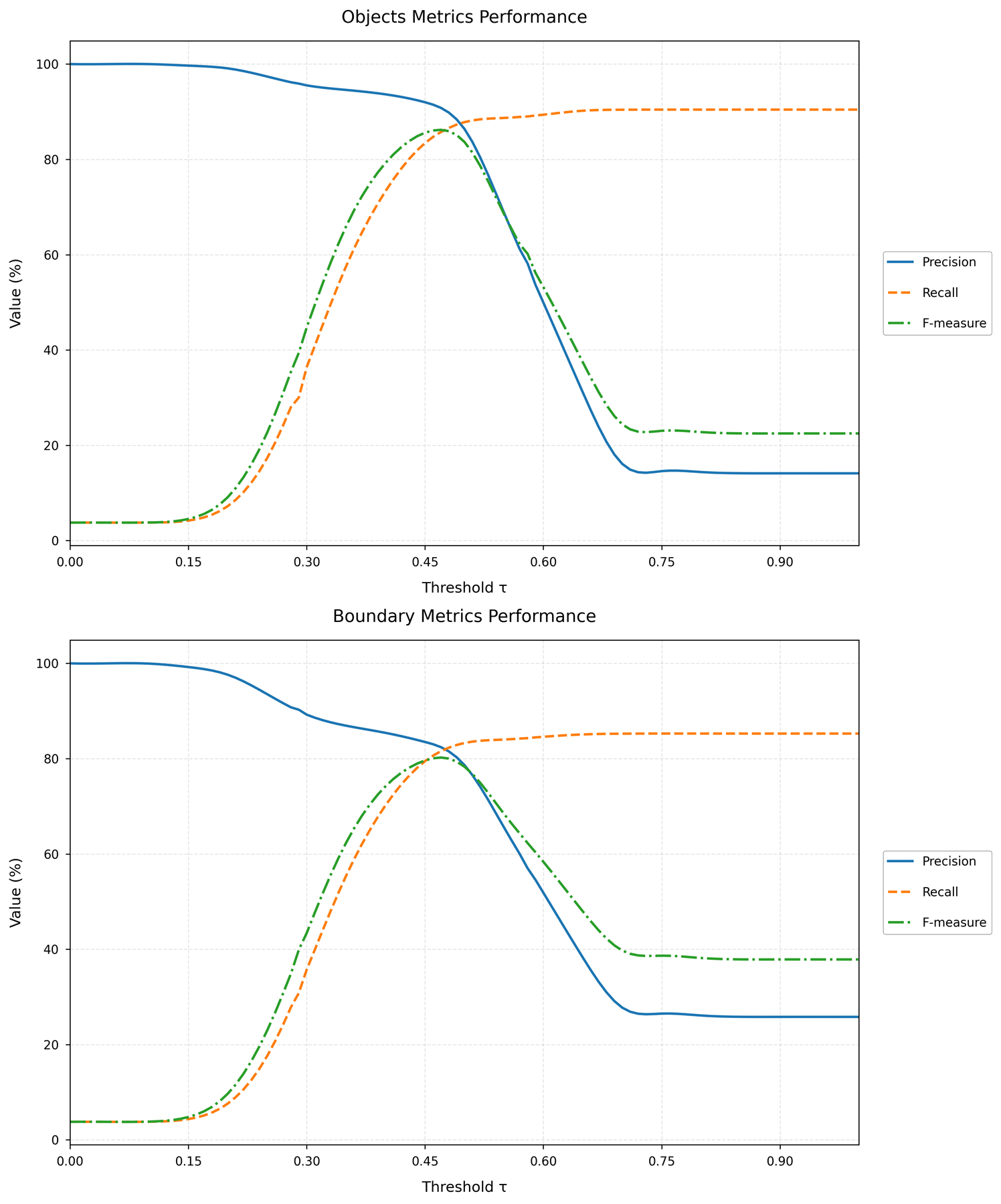}
    \centering
    \caption{Ablation studies examining the sensitivity of model performance metrics relative to threshold parameter $\tau$ for both object and boundary detection.}
    \label{fig9}
    \end{figure}
The background similarity threshold $\tau$ serves as a critical parameter in our method, determining the classification of predictions as either foreground objects or background regions. A predicted mask is classified as background only when its average background similarity exceeds $\tau$. Fig. \ref{fig9} presents ablation studies on both object and boundary metrics, illustrating the model's sensitivity to variations in $\tau$. These experiments, conducted on the OCID dataset without point prompts, reveal distinct performance patterns across different threshold ranges. At low thresholds ($\tau < 0.15$), all metrics maintain relatively stable but low values, indicating conservative background classification with only highly confident detections retained. As $\tau$ increases from 0.15 to 0.47, both recall and F-measure increase dramatically, while precision decreases gradually from its maximum value. This indicates that our background mask filtering step based on explicit visual representations is effective, as it filters out masks with a similarity to background patches exceeding a threshold $\tau$, while preserving foreground object masks. 
The F-measure, which harmonically balances precision and recall, reaches its optimal value at $\tau \approx 0.47$. Beyond this point, the precision decreases dramatically while recall increases marginally, plateauing around 90\%. This behavior suggests that higher thresholds lead to excessive prediction acceptance, significantly increasing false positives while only marginally improving true positive detection. Both object and boundary metrics exhibit similar trends, though boundary detection shows slightly lower overall performance, reflecting the increased complexity of precise boundary delineation. The convergence of metrics at high thresholds ($\tau > 0.75$) indicates that the computed background patches $l$ have limited similarity to other image patches, resulting in fewer background classifications as $\tau$ increases.

\subsubsection{Prompt Ablation}
\begin{table}
    \centering
    \caption{The ablation study results of different prompt methods on the OCID dataset \cite{suchi2019easylabel}.}
    \begin{tabular}{l|ccc|ccc}
        \hline
                                 & \multicolumn{3}{c|}{Overlap}                                                   & \multicolumn{3}{c}{Boundary}                                                   \\
        \multirow{-2}{*}{Prompt} & {\color[HTML]{CE6301} P} & {\color[HTML]{329A9D} R} & {\color[HTML]{9A0000} F} & {\color[HTML]{CE6301} P} & {\color[HTML]{329A9D} R} & {\color[HTML]{9A0000} F} \\ \hline
        Boxes                    & 90.5                     & 84.6                    & 87.4                     & 83.6                     & 79.4                    & 81.5                     \\
        Random points            & 91.9                     & 85.8                     & 88.6                     & 84.4                     & 81.7                    & 82.8                     \\
        Cluster centers           & 92.5                     & 86.1                     & 89.2                     & 84.5                     & 82.4                     & 83.5                     \\ \hline
        \end{tabular}
        \label{table5}
        \end{table}
Table \ref{table5} presents the experimental results of various prompt methods on the OCID dataset. Among these, the point-based method surpasses the box-based method in performance. This is likely due to the presence of overlapping objects in the benchmark dataset, which makes it challenging for box-based methods to accurately locate each object, whereas point-based methods handle this issue more effectively. Within the point selection techniques, the results of Cluster and Random are comparable, with Cluster having a slight edge. This advantage is primarily because the K-Medoids cluster algorithm ensures good coverage of different parts of the objects and exhibits robustness against noise and outliers.

\subsection{Discussion of Failure Cases} \label{section45}
\begin{figure}
	\includegraphics[width=3.4in]{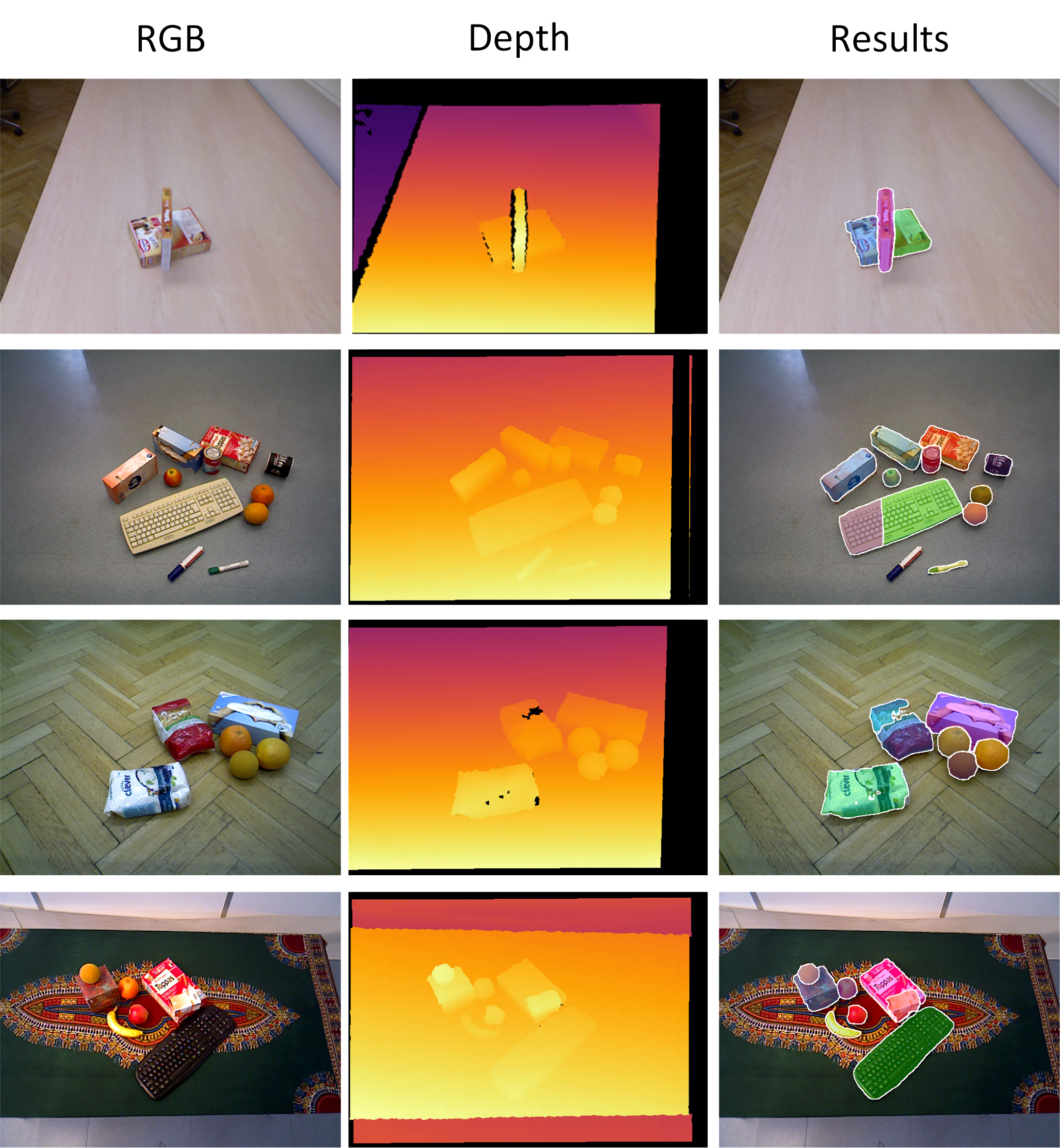}
	\centering
	\caption{Common failure cases of our proposed method in OCID\cite{suchi2019easylabel} and OSD\cite{richtsfeld2012segmentation} datasets. The segmentation results are not refined using point prompts.}
	\label{fig7}
\end{figure}

Fig. \ref{fig7} illustrates the typical failure modes of our proposed method. The first row illustrates over-segmentation of occluded objects when placed in close proximity. Additionally, this issue recurs with elongated objects, as shown by the keyboard in the second row, where a significant variation in object depth across its length leads to over-segmentation. The second row also highlights a scenario of under-segmentation, exemplified by a pen that remains unsegmented. This may be due to the increasing complexity of the scene, which causes the attention capacity of the self-supervised ViT to distribute across multiple regions. Consequently, objects with less prominent features receive lower attention weights. In the third row, the segmentation results illustrate that our method relies on high-quality depth images to produce the initial mask proposal. With a structured light depth camera, reflections from certain materials (e.g., reflections from the surface of a bag) can impair the depth image quality and affect the segmentation accuracy. The fourth row discusses challenges in segmenting stacked flat-surfaced objects, such as cereal boxes, from depth maps alone due to their presentation as a single flat surface. Future work could explore enhancing ViT features to reduce over-segmentation and employing multi-view images to address under-segmentation.

\subsection{Testing with a Real Robot} \label{section46}
\begin{figure}
	\includegraphics[width=3.4in]{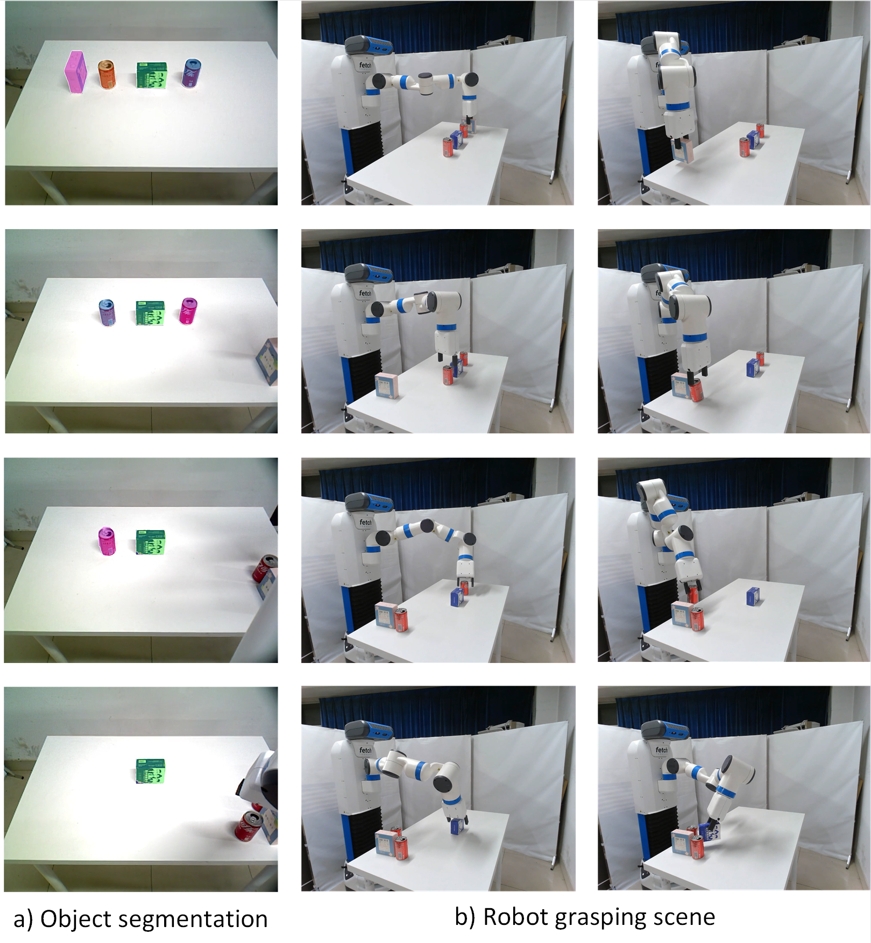}
	\centering
	\caption{Demonstrating manipulation of unknown objects in the cluttered environment with a fetch robot equipped with a two-finger gripper. (Please see supplementary video.)}
	\label{fig6}
\end{figure}

The applicability of our proposed segmentation approach is validated through real-world robotic manipulation experiments using the Fetch mobile robot platform. The experimental setup employs the robot's head-mounted RGB-D camera operating at 640×480 resolution for scene perception. The implementation pipeline integrates three key components: our proposed segmentation network for object  segmentation, Contact-GraspNet \cite{sundermeyer2021contact} for grasp pose synthesis, and MoveIt \cite{zhang2021safe, chitta2012moveit} for motion planning and execution. During manipulation, RGB-D images captured by the robot's camera are processed through our segmentation network to generate precise object-wise masks. These segmentation results are then utilized by Contact-GraspNet to compute feasible grasp poses. Subsequently, MoveIt generates collision-free trajectories while accounting for the robot's kinematic constraints. Fig. \ref{fig6} illustrates the segmentation and grasping process through various stages, culminating in the robot's successful execution of object grasping.

\section{Conclusion} \label{section5}
This paper presented ZISVFM, a novel and effective approach for zero-shot object segmentation in robotic applications. The ZISVFM methodology comprised three distinct stages. In the first stage, the system generated object-agnostic mask proposals by converting depth images to colorized representations using the viridis color map, which were subsequently processed through SAM to generate diverse masks. The second stage employed a self-supervised ViT to refine these proposals through explicit visual representations, utilizing attention maps and weighted feature representations to identify and eliminate non-object masks. In the final stage, the refined masks served as foundations for object localization, where K-Medoids clustering identified key points to guide SAM in achieving precise segmentation.
Experimental results on multiple benchmark datasets in desktop environments, along with tests on a self-collected
dataset featuring complex, multi-level scenarios, demonstrated the superior performance and practicality of ZISVFM. Future work will focus on further reducing the sim2real gap by integrating more robust domain adaptation techniques and exploring the multi-view images for mitigating object clutter and occlusion to enhance the adaptability and efficiency of our model.

\bibliographystyle{IEEEtran}
\bibliography{mybibfile}

\begin{thebibliography}{10}
\providecommand{\url}[1]{#1}
\csname url@samestyle\endcsname
\providecommand{\newblock}{\relax}
\providecommand{\bibinfo}[2]{#2}
\providecommand{\BIBentrySTDinterwordspacing}{\spaceskip=0pt\relax}
\providecommand{\BIBentryALTinterwordstretchfactor}{4}
\providecommand{\BIBentryALTinterwordspacing}{\spaceskip=\fontdimen2\font plus
\BIBentryALTinterwordstretchfactor\fontdimen3\font minus
  \fontdimen4\font\relax}
\providecommand{\BIBforeignlanguage}[2]{{%
\expandafter\ifx\csname l@#1\endcsname\relax
\typeout{** WARNING: IEEEtran.bst: No hyphenation pattern has been}%
\typeout{** loaded for the language `#1'. Using the pattern for}%
\typeout{** the default language instead.}%
\else
\language=\csname l@#1\endcsname
\fi
#2}}
\providecommand{\BIBdecl}{\relax}
\BIBdecl

\bibitem{xie2021unseen}
C.~Xie, Y.~Xiang, A.~Mousavian, and D.~Fox, ``Unseen object instance
  segmentation for robotic environments,'' \emph{IEEE Transactions on
  Robotics}, vol.~37, no.~5, pp. 1343--1359, 2021.

\bibitem{fu2024taylor}
K.~Fu, X.~Dang, and Y.~Zhang, ``Taylor neural network for unseen object
  instance segmentation in hierarchical grasping,'' \emph{IEEE/ASME
  Transactions on Mechatronics}, 2024, to be published, DOI:
  10.1109/TMECH.2023.3347558.

\bibitem{richtsfeld2012segmentation}
A.~Richtsfeld, T.~M{\"o}rwald, J.~Prankl, M.~Zillich, and M.~Vincze,
  ``Segmentation of unknown objects in indoor environments,'' in \emph{2012
  IEEE/RSJ International Conference on Intelligent Robots and Systems}, 2012,
  pp. 4791--4796.

\bibitem{ceola2022learn}
F.~Ceola, E.~Maiettini, G.~Pasquale, G.~Meanti, L.~Rosasco, and L.~Natale,
  ``Learn fast, segment well: Fast object segmentation learning on the icub
  robot,'' \emph{IEEE Transactions on Robotics}, vol.~38, no.~5, pp.
  3154--3172, 2022.

\bibitem{xiang2021learning}
Y.~Xiang, C.~Xie, A.~Mousavian, and D.~Fox, ``Learning rgb-d feature embeddings
  for unseen object instance segmentation,'' in \emph{Conference on Robot
  Learning}, 2021, pp. 461--470.

\bibitem{li2023stow}
Y.~Li, M.~Zhang, M.~Grotz, K.~Mo, and D.~Fox, ``Stow: Discrete-frame
  segmentation and tracking of unseen objects for warehouse picking robots,''
  in \emph{Proceedings of the Conference on Robot Learning (CoRL)}, 2023.

\bibitem{he2017mask}
K.~He, G.~Gkioxari, P.~Doll{\'a}r, and R.~Girshick, ``Mask r-cnn,'' in
  \emph{Proceedings of the IEEE International Conference on Computer Vision},
  2017, pp. 2961--2969.

\bibitem{minaee2021image}
S.~Minaee, Y.~Boykov, F.~Porikli, A.~Plaza, N.~Kehtarnavaz, and D.~Terzopoulos,
  ``Image segmentation using deep learning: A survey,'' \emph{IEEE Transactions
  on Pattern Analysis and Machine Intelligence}, vol.~44, no.~7, pp.
  3523--3542, 2022.

\bibitem{russakovsky2015imagenet}
O.~Russakovsky, J.~Deng, H.~Su, J.~Krause, S.~Satheesh, S.~Ma, Z.~Huang,
  A.~Karpathy, A.~Khosla, M.~Bernstein \emph{et~al.}, ``Imagenet large scale
  visual recognition challenge,'' \emph{International Journal of Computer
  Vision}, vol. 115, pp. 211--252, 2015.

\bibitem{lin2014microsoft}
T.-Y. Lin, M.~Maire, S.~Belongie, J.~Hays, P.~Perona, D.~Ramanan,
  P.~Doll{\'a}r, and C.~L. Zitnick, ``Microsoft coco: Common objects in
  context,'' in \emph{European Conference Computer Vision}, 2014, pp. 740--755.

\bibitem{back2022unseen}
S.~Back, J.~Lee, T.~Kim, S.~Noh, R.~Kang, S.~Bak, and K.~Lee, ``Unseen object
  amodal instance segmentation via hierarchical occlusion modeling,'' in
  \emph{2022 International Conference on Robotics and Automation (ICRA)}, 2022,
  pp. 5085--5092.

\bibitem{9636281}
M.~Durner, W.~Boerdijk, M.~Sundermeyer, W.~Friedl, Z.-C. Márton, and
  R.~Triebel, ``Unknown object segmentation from stereo images,'' in \emph{2021
  IEEE/RSJ International Conference on Intelligent Robots and Systems (IROS)},
  2021, pp. 4823--4830.

\bibitem{xie2020best}
C.~Xie, Y.~Xiang, A.~Mousavian, and D.~Fox, ``The best of both modes:
  Separately leveraging rgb and depth for unseen object instance
  segmentation,'' in \emph{Conference on Robot Learning}, 2020, pp. 1369--1378.

\bibitem{danielczuk2019segmenting}
M.~Danielczuk, M.~Matl, S.~Gupta, A.~Li, A.~Lee, J.~Mahler, and K.~Goldberg,
  ``Segmenting unknown 3d objects from real depth images using mask r-cnn
  trained on synthetic data,'' in \emph{2019 International Conference on
  Robotics and Automation (ICRA)}, 2019, pp. 7283--7290.

\bibitem{zhang2023unseen}
L.~Zhang, S.~Zhang, X.~Yang, H.~Qiao, and Z.~Liu, ``Unseen object instance
  segmentation with fully test-time rgb-d embeddings adaptation,'' in
  \emph{2023 IEEE International Conference on Robotics and Automation (ICRA)},
  2023, pp. 4945--4952.

\bibitem{lu2023self}
Y.~Lu, N.~Khargonkar, Z.~Xu, C.~Averill, K.~Palanisamy, K.~Hang, Y.~Guo,
  N.~Ruozzi, and Y.~Xiang, ``Self-supervised unseen object instance
  segmentation via long-term robot interaction,'' \emph{arXiv preprint
  arXiv:2302.03793}, 2023.

\bibitem{yu2022self}
H.~Yu and C.~Choi, ``Self-supervised interactive object segmentation through a
  singulation-and-grasping approach,'' in \emph{European Conference on Computer
  Vision}, 2022, pp. 621--637.

\bibitem{eitel2019self}
A.~Eitel, N.~Hauff, and W.~Burgard, ``Self-supervised transfer learning for
  instance segmentation through physical interaction,'' in \emph{2019 IEEE/RSJ
  International Conference on Intelligent Robots and Systems (IROS)}, 2019, pp.
  4020--4026.

\bibitem{kirillov2023segment}
A.~Kirillov, E.~Mintun, N.~Ravi, H.~Mao, C.~Rolland, L.~Gustafson, T.~Xiao,
  S.~Whitehead, A.~C. Berg, W.-Y. Lo \emph{et~al.}, ``Segment anything,''
  \emph{arXiv preprint arXiv:2304.02643}, 2023.

\bibitem{wang2023seggpt}
X.~Wang, X.~Zhang, Y.~Cao, W.~Wang, C.~Shen, and T.~Huang, ``Seggpt: Segmenting
  everything in context,'' in \emph{Proceedings of the IEEE/CVF International
  Conference on Computer Vision (ICCV)}, 2023, pp. 1130--1140.

\bibitem{chen2023semantic}
J.~Chen, Z.~Yang, and L.~Zhang, ``Semantic segment anything,''
  \url{https://github.com/fudan-zvg/Semantic-Segment-Anything}, 2023.

\bibitem{zou2023segment}
X.~Zou, J.~Yang, H.~Zhang, F.~Li, L.~Li, J.~Gao, and Y.~J. Lee, ``Segment
  everything everywhere all at once,'' in \emph{Proceedings of the 37th
  International Conference on Neural Information Processing Systems}, 2023, pp.
  19\,769--19\,782.

\bibitem{hunter2007matplotlib}
J.~D. Hunter, ``Matplotlib: A 2d graphics environment,'' \emph{Computing in
  science \& engineering}, vol.~9, no.~03, pp. 90--95, 2007.

\bibitem{oquab2023dinov2}
M.~Oquab, T.~Darcet, T.~Moutakanni, H.~Vo, M.~Szafraniec, V.~Khalidov,
  P.~Fernandez, D.~Haziza, F.~Massa, A.~El-Nouby \emph{et~al.}, ``Dinov2:
  Learning robust visual features without supervision,'' \emph{Transactions on
  Machine Learning Research}, 2024.

\bibitem{xie2022rice}
C.~Xie, A.~Mousavian, Y.~Xiang, and D.~Fox, ``Rice: Refining instance masks in
  cluttered environments with graph neural networks,'' in \emph{Conference on
  Robot Learning}, 2022, pp. 1655--1665.

\bibitem{chen20233d}
J.~Chen, M.~Sun, T.~Bao, R.~Zhao, L.~Wu, and Z.~He, ``3d model-based zero-shot
  pose estimation pipeline,'' \emph{arXiv preprint arXiv:2305.17934}, 2023.

\bibitem{nguyen2023cnos}
V.~N. Nguyen, T.~Groueix, G.~Ponimatkin, V.~Lepetit, and T.~Hodan, ``Cnos: A
  strong baseline for cad-based novel object segmentation,'' in
  \emph{Proceedings of the IEEE/CVF International Conference on Computer
  Vision}, 2023, pp. 2134--2140.

\bibitem{kirillov2019panoptic}
A.~Kirillov, K.~He, R.~Girshick, C.~Rother, and P.~Doll{\'a}r, ``Panoptic
  segmentation,'' in \emph{Proceedings of the IEEE/CVF Conference on Computer
  Vision and Pattern Pecognition}, 2019, pp. 9404--9413.

\bibitem{bolya2019yolact}
D.~Bolya, C.~Zhou, F.~Xiao, and Y.~J. Lee, ``Yolact: Real-time instance
  segmentation,'' in \emph{Proceedings of the IEEE/CVF International Conference
  on Computer Vision}, 2019, pp. 9157--9166.

\bibitem{long2015fully}
J.~Long, E.~Shelhamer, and T.~Darrell, ``Fully convolutional networks for
  semantic segmentation,'' in \emph{Proceedings of the IEEE Conference on
  Computer Vision and Pattern Recognition}, 2015, pp. 3431--3440.

\bibitem{awais2023foundational}
M.~Awais, M.~Naseer, S.~Khan, R.~M. Anwer, H.~Cholakkal, M.~Shah, M.-H. Yang,
  and F.~S. Khan, ``Foundational models defining a new era in vision: A survey
  and outlook,'' \emph{arXiv preprint arXiv:2307.13721}, 2023.

\bibitem{he2023accuracy}
S.~He, R.~Bao, J.~Li, P.~E. Grant, and Y.~Ou, ``Accuracy of segment-anything
  model (sam) in medical image segmentation tasks,'' \emph{arXiv preprint
  arXiv:2304.09324}, 2023.

\bibitem{cheng2023segment}
Y.~Cheng, L.~Li, Y.~Xu, X.~Li, Z.~Yang, W.~Wang, and Y.~Yang, ``Segment and
  track anything,'' \emph{arXiv preprint arXiv:2305.06558}, 2023.

\bibitem{yu2023inpaint}
T.~Yu, R.~Feng, R.~Feng, J.~Liu, X.~Jin, W.~Zeng, and Z.~Chen, ``Inpaint
  anything: Segment anything meets image inpainting,'' \emph{arXiv preprint
  arXiv:2304.06790}, 2023.

\bibitem{cui2023all}
C.~Cui, R.~Deng, Q.~Liu, T.~Yao, S.~Bao, L.~W. Remedios, Y.~Tang, and Y.~Huo,
  ``All-in-sam: from weak annotation to pixel-wise nuclei segmentation with
  prompt-based finetuning,'' \emph{Journal of Physics: Conference Series}, vol.
  2722, no.~1, p. 012012, 2024.

\bibitem{zhang2023uvosam}
Z.~Zhang, Z.~Wei, S.~Zhang, Z.~Dai, and S.~Zhu, ``Uvosam: A mask-free paradigm
  for unsupervised video object segmentation via segment anything model,''
  \emph{arXiv preprint arXiv:2305.12659}, 2023.

\bibitem{vaswani2017attention}
A.~Vaswani, N.~Shazeer, N.~Parmar, J.~Uszkoreit, L.~Jones, A.~N. Gomez,
  {\L}.~Kaiser, and I.~Polosukhin, ``Attention is all you need,'' in
  \emph{Proceedings of the 31st International Conference on Neural Information
  Processing Systems}, 2017, pp. 6000--6010.

\bibitem{dosovitskiy2020image}
A.~Dosovitskiy, L.~Beyer, A.~Kolesnikov, D.~Weissenborn, X.~Zhai,
  T.~Unterthiner, M.~Dehghani, M.~Minderer, G.~Heigold, S.~Gelly \emph{et~al.},
  ``An image is worth 16x16 words: Transformers for image recognition at
  scale,'' in \emph{9th International Conference on Learning Representations},
  2021.

\bibitem{zhang2023cross}
Y.~Zhang, M.~Yin, H.~Wang, and C.~Hua, ``Cross-level multi-modal features
  learning with transformer for rgb-d object recognition,'' \emph{IEEE
  Transactions on Circuits and Systems for Video Technology}, vol.~33, no.~12,
  pp. 7121--7130, 2023.

\bibitem{chen2021mocov3}
X.~Chen, S.~Xie, and K.~He, ``An empirical study of training self-supervised
  vision transformers,'' in \emph{Proceedings of the IEEE/CVF International
  Conference on Computer Vision (ICCV)}, 2021, pp. 9640--9649.

\bibitem{caron2021emerging}
M.~Caron, H.~Touvron, I.~Misra, H.~J{\'e}gou, J.~Mairal, P.~Bojanowski, and
  A.~Joulin, ``Emerging properties in self-supervised vision transformers,'' in
  \emph{Proceedings of the IEEE/CVF international conference on computer
  vision}, 2021, pp. 9650--9660.

\bibitem{li2021mst}
Z.~Li, Z.~Chen, F.~Yang, W.~Li, Y.~Zhu, C.~Zhao, R.~Deng, L.~Wu, R.~Zhao,
  M.~Tang \emph{et~al.}, ``Mst: Masked self-supervised transformer for visual
  representation,'' \emph{Advances in Neural Information Processing Systems},
  vol.~34, pp. 13\,165--13\,176, 2021.

\bibitem{devlin2018bert}
J.~Devlin, M.-W. Chang, K.~Lee, and K.~Toutanova, ``Bert: Pre-training of deep
  bidirectional transformers for language understanding,'' in \emph{Proceedings
  of the 2019 Conference of the North American Chapter of the Association for
  Computational Linguistics: Human Language Technologies (NAACL-HLT)}, 2019,
  pp. 4171--4186.

\bibitem{bao2021beit}
H.~Bao, L.~Dong, S.~Piao, and F.~Wei, ``Beit: Bert pre-training of image
  transformers,'' in \emph{The International Conference on Learning
  Representations (ICLR)}, 2022.

\bibitem{he2022masked}
K.~He, X.~Chen, S.~Xie, Y.~Li, P.~Doll{\'a}r, and R.~Girshick, ``Masked
  autoencoders are scalable vision learners,'' in \emph{Proceedings of the
  IEEE/CVF Conference on Computer Vision and Pattern Recognition}, 2022, pp.
  16\,000--16\,009.

\bibitem{amir2021deep}
S.~Amir, Y.~Gandelsman, S.~Bagon, and T.~Dekel, ``Deep vit features as dense
  visual descriptors,'' in \emph{European Conference on Computer Vision},
  vol.~2, no.~3, 2022, p.~4.

\bibitem{suchi2019easylabel}
M.~Suchi, T.~Patten, D.~Fischinger, and M.~Vincze, ``Easylabel: A
  semi-automatic pixel-wise object annotation tool for creating robotic rgb-d
  datasets,'' in \emph{2019 International Conference on Robotics and Automation
  (ICRA)}, 2019, pp. 6678--6684.

\bibitem{park2009simple}
H.-S. Park and C.-H. Jun, ``A simple and fast algorithm for k-medoids
  clustering,'' \emph{Expert Systems With Applications}, vol.~36, no.~2, pp.
  3336--3341, 2009.

\bibitem{darcet2023vitneedreg}
T.~Darcet, M.~Oquab, J.~Mairal, and P.~Bojanowski, ``Vision transformers need
  registers,'' in \emph{The Twelfth International Conference on Learning
  Representations (ICLR)}, 2024.

\bibitem{paszke2019pytorch}
A.~Paszke, S.~Gross, F.~Massa, A.~Lerer, J.~Bradbury, G.~Chanan, T.~Killeen,
  Z.~Lin, N.~Gimelshein, L.~Antiga \emph{et~al.}, ``Pytorch: An imperative
  style, high-performance deep learning library,'' \emph{Advances in neural
  information processing systems}, vol.~32, 2019.

\bibitem{Dave_2019_ICCV}
A.~Dave, P.~Tokmakov, and D.~Ramanan, ``Towards segmenting anything that
  moves,'' in \emph{Proceedings of the IEEE/CVF International Conference on
  Computer Vision (ICCV) Workshops}, Oct 2019.

\bibitem{lu2022mean}
Y.~Lu, Y.~Chen, N.~Ruozzi, and Y.~Xiang, ``Mean shift mask transformer for
  unseen object instance segmentation,'' \emph{arXiv preprint
  arXiv:2211.11679}, 2022.

\bibitem{sundermeyer2021contact}
M.~Sundermeyer, A.~Mousavian, R.~Triebel, and D.~Fox, ``Contact-graspnet:
  Efficient 6-dof grasp generation in cluttered scenes,'' in \emph{2021 IEEE
  International Conference on Robotics and Automation (ICRA)}, 2021, pp.
  13\,438--13\,444.

\bibitem{zhang2021safe}
Y.~Zhang, G.~Tian, and X.~Shao, ``Safe and efficient robot manipulation:
  Task-oriented environment modeling and object pose estimation,'' \emph{IEEE
  Transactions on Instrumentation and Measurement}, vol.~70, pp. 1--12, 2021.

\bibitem{chitta2012moveit}
S.~Chitta, I.~Sucan, and S.~Cousins, ``Moveit![ros topics],'' \emph{IEEE
  Robotics \& Automation Magazine}, vol.~19, no.~1, pp. 18--19, 2012.

\end{thebibliography}

\begin{IEEEbiography}[{\includegraphics[width=1in,height=1.25in,clip,keepaspectratio]{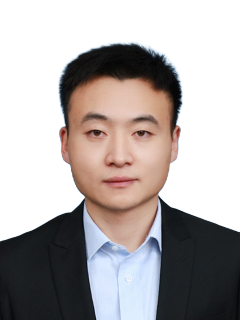}}]{Ying Zhang} (Senior Member, IEEE) received the Ph.D. degree in Control Theory and Control Engineering from Shandong University, Jinan, China, in 2021.

He is currently an Associate Professor with the School of Electrical Engineering, Yanshan University. He is an Early Career Associate Editor for \emph{Biomimetic Intelligence and Robotics}, \emph{Intelligence \& Robotics}, and \emph{Brain-X}. Dr. Zhang received Best Paper Award in CCIR 2018, the Youth Top Talent Project for Hebei Education Department in 2023, and the Excellent Youth Project for Hebei NSF in 2024. His current research interests include intelligent robot systems, visual perception, environment modeling, robot skill learning.
\end{IEEEbiography}

\begin{IEEEbiography}[{\includegraphics[width=1in,height=1.25in,clip,keepaspectratio]{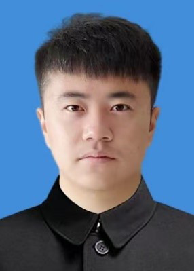}}]{Maoliang Yin} received the B.S. degree in building electrical and intelligent engineering from Qingdao University of Technology, Qingdao, China, in 2021. He is currently working toward the Ph.D. degree in control science and engineering with Yanshan University, Qinhuangdao, China.

His current research interests include service robots, deep learning, and computer vision.
\end{IEEEbiography}

\begin{IEEEbiography}[{\includegraphics[width=1in,height=1.25in,clip,keepaspectratio]{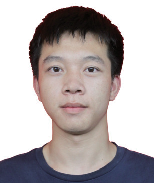}}]{Wenfu Bi} received the B.S. degree in School of Information and Electrical Engineering from the Hunan University of Science and Technology, Xiangtan, China, in 2021. He is currently working toward the M.S. degree in control science and engineering at Yanshan University, Qinhuangdao, China.

His research interests include mapping, localization and navigation of mobile robots, environmental modeling.
\end{IEEEbiography}

\begin{IEEEbiography}[{\includegraphics[width=1in,height=1.25in,clip,keepaspectratio]{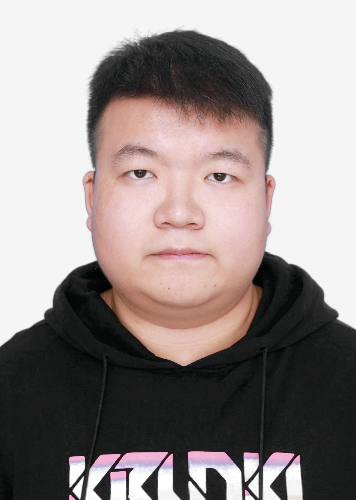}}]{Haibao Yan} received the B.S. degree in automation from Taiyuan University of Technology, Taiyuan, China, in 2021. He is currently pursuing the M.S. degree in control engineering at Yanshan University, Qinhuangdao, China. 

His current research interests include air-ground collaborative robots and simultaneous localization and mapping (SLAM).
\end{IEEEbiography}

\begin{IEEEbiography}[{\includegraphics[width=1in,height=1.25in,clip,keepaspectratio]{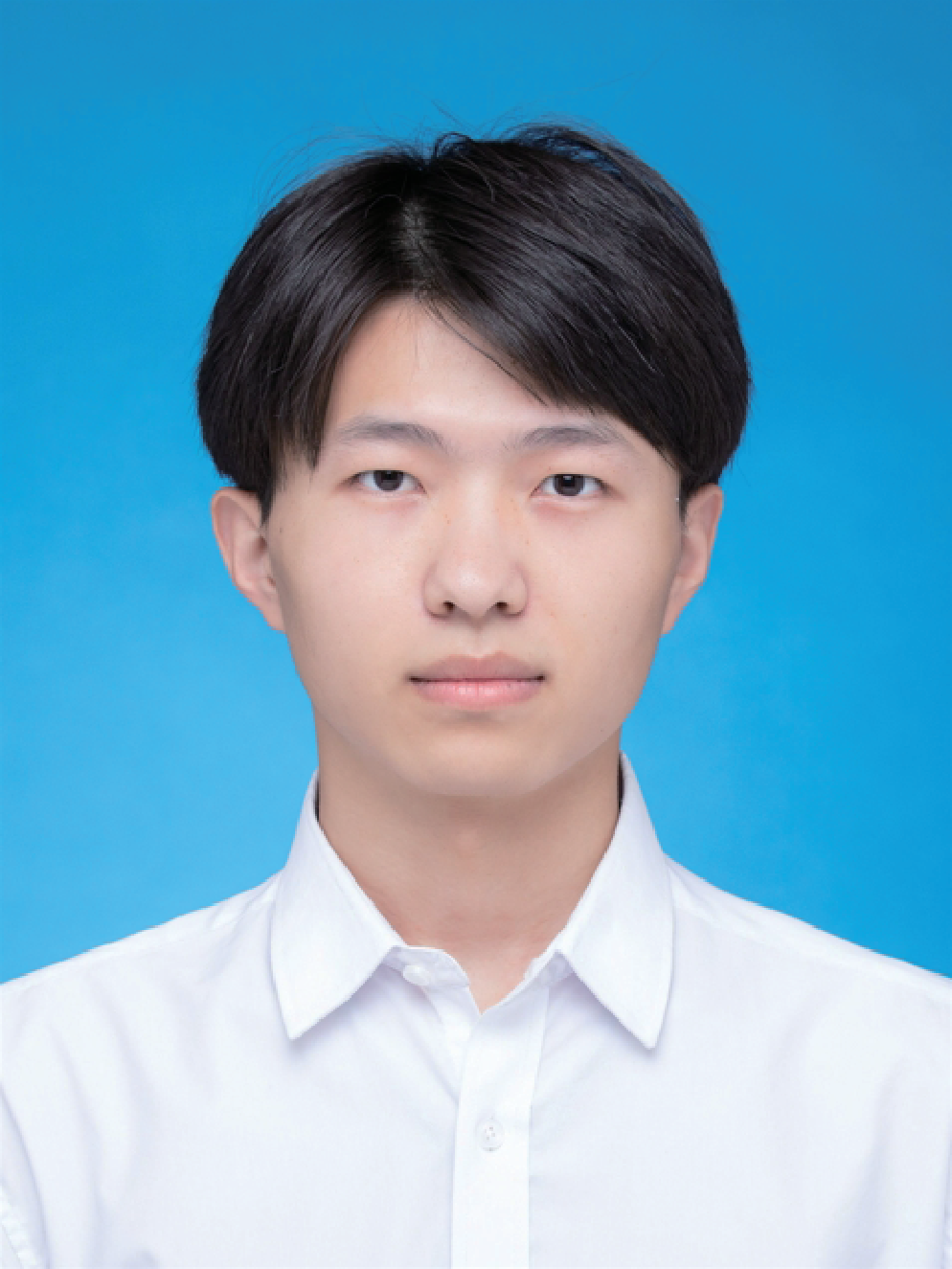}}]{Shaohan Bian} received the B.S. degree from Jiangsu University, Zhenjiang, China, in 2023. He is currently working toward the M.S. degree in control engineering at Yanshan University, Qinhuangdao, China.

His current research interests include service robots and task planing.
\end{IEEEbiography}

\begin{IEEEbiography}[{\includegraphics[width=1in,height=1.25in,clip,keepaspectratio]{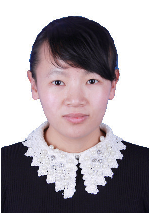}}]{Cui-Hua Zhang} (Member, IEEE)
received the M.S. degree in control theory and control engineering from the Institute of Automation, Qufu Normal University, Qufu, China, in 2017, and the Ph.D. degree in control theory and control engineering from Northeastern University, Shenyang, China, in 2021. 

She is currently an Associate Professor with the School of Electrical Engineering, Yanshan University, Qinghuangdao, China. Her current research interests include intelligent robot systems, nonlinear adaptive control and networked nonlinear systems.
\end{IEEEbiography}

\begin{IEEEbiography}[{\includegraphics[width=1in,height=1.25in,clip,keepaspectratio]{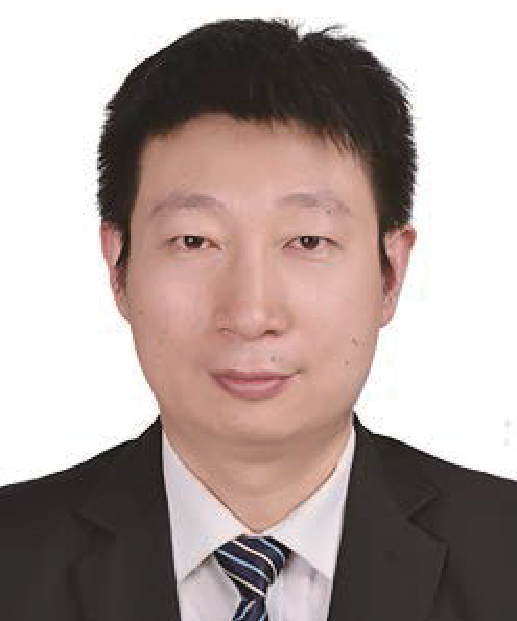}}]{Changchun Hua} (Fellow, IEEE) received the Ph.D. degree in Electrical Engineering from Yanshan University, Qinhuangdao, China, in 2005. From 2006 to 2007, he was a Research Fellow with the National University of Singapore, Singapore. From 2007 to 2009, he was with Carleton University, Ottawa, ON, Canada, funded by the Province of Ontario Ministry of Research and Innovation Program. From 2009 to 2010, he was with the University of Duisburg-Essen, Essen, Germany, funded by the Alexander von Humboldt Foundation.

He is currently a Full Professor with Yanshan University. He has been involved in more than 15 projects supported by the National Natural Science Foundation of China, the National Education Committee Foundation of China, and other important foundations. He is a Cheung Kong Scholars Programme Special Appointment Professor. He is an Associate Editor for the \textsc{IEEE Transactions on Cybernetics}, \emph{International Journal of Control, Automation and Systems}, and other journals. He is a highly cited Researcher (since 2014) selected by Elsevier. His research interests include nonlinear control systems, multiagent systems, teleoperation systems, and intelligent robot systems.
\end{IEEEbiography}

\end{document}